\documentclass[a4paper,fleqn]{cas-sc}

\usepackage[numbers]{natbib}
\usepackage{times}
\usepackage{soul}
\usepackage{url}

\usepackage[utf8]{inputenc}
\usepackage[small]{caption}
\usepackage{graphicx}
\usepackage{amsmath}
\usepackage{amsthm}
\usepackage{booktabs}
\usepackage{algorithm}
\usepackage{algorithmic}
\usepackage[switch]{lineno}
\newcommand{\indep}{\perp \!\!\! \perp}

\usepackage[utf8]{inputenc} % allow utf-8 input
\usepackage[T1]{fontenc}    % use 8-bit T1 fonts
\usepackage{url}            % simple URL typesetting
\usepackage{booktabs}       % professional-quality tables
\usepackage{amsfonts}       % blackboard math symbols
\usepackage{nicefrac}       % compact symbols for 1/2, etc.
\usepackage{microtype}      % microtypography
\usepackage{xcolor} % 加载xcolor，指定选项
\usepackage{colortbl} % 现在不会冲突，因为xcolor已正确配置
\usepackage{wrapfig,lipsum,booktabs}

\usepackage{algorithm}
\usepackage{algorithmic}
\usepackage{subcaption}
\usepackage{epstopdf}
\usepackage{soul}
\usepackage{multicol}
\usepackage{multirow}
\usepackage{float}
\usepackage{enumitem}
\usepackage{booktabs}
\usepackage{amsmath}
\usepackage{amsthm}
\usepackage{amssymb}
\usepackage{bm}
\usepackage{booktabs}
\usepackage{caption}
\usepackage{hyperref}

\definecolor{lightblue}{rgb}{0.93,0.95,1.0}

\newtheorem{proposition}{Proposition}
\usepackage[normalem]{ulem}
\useunder{\uline}{\ul}{}
\newcommand{\tabincell}[2]{\begin{tabular}{@{}#1@{}}#2\end{tabular}}

\begin{document}
\let\WriteBookmarks\relax
\def\floatpagepagefraction{1}
\def\textpagefraction{.001}

\shorttitle{}    

\shortauthors{S. Tao, Q. Cao, H. Shen et al.}  

\title [mode = title]{IDEA: Invariant Defense for Graph Adversarial Robustness}  

%\tnotemark[<tnote number>] 

%\tnotetext[<tnote number>]{<tnote text>} 


\author[1,2]{Shuchang Tao}[orcid=0000-0001-6113-6145]

%\cormark[<corr mark no>]

%\fnmark[<footnote mark no>]

\ead{shuchangtao5@gmail.com, tshuchang@163.com}

%\ead[url]{<URL>}

\credit{Adversarial attack, Graph neural networks, Data mining}

\author[1]{Qi Cao}
% Footnote of the second author

% Email id of the second author
\ead{caoqi@ict.ac.cn}


\credit{}
\cormark[1]

\author[1,2]{Huawei Shen}
% Footnote of the second author
%\fnmark[3]

% Email id of the second author
\ead{shenhuawei@ict.ac.cn}


\credit{}
\cormark[1]

\author[1,2]{Yunfan Wu}
% Footnote of the second author
%\fnmark[3]

% Email id of the second author
\ead{wuyunfan19b@ict.ac.cn}


\credit{}

\author[1]{Bingbing Xu}
% Footnote of the second author
%\fnmark[3]

% Email id of the second author
\ead{xubingbing@ict.ac.cn}


\credit{}

\author[1,2]{Xueqi Cheng}
% Footnote of the second author
%\fnmark[3]
\ead{cxq@ict.ac.cn}

\credit{}

\affiliation[1]{organization={CAS Key Laboratory of AI Safety,
Institute of Computing Technology,
Chinese Academy of Sciences},
            city={Beijing},
%          citysep={}, % Uncomment if no comma needed between city and postcode
            country={China}}
            
%\affiliation[2]{organization={CAS Key Laboratory of Network Data Science and Technology, Institute of Computing Technology, Chinese Academy of Sciences},
%            city={Beijing},
%%          citysep={}, % Uncomment if no comma needed between city and postcode
%            country={China}}
            
\affiliation[2]{organization={University of Chinese Academy of Sciences},
            city={Beijing},
            country={China}}

\cortext[1]{Corresponding author}



\begin{abstract}
Despite the success of graph neural networks (GNNs), their vulnerability to adversarial attacks poses tremendous challenges for practical applications. Existing defense methods suffer from severe performance decline under unseen attacks, due to either limited observed adversarial examples or pre-defined heuristics. To address these limitations, we analyze the causalities in graph adversarial attacks and conclude that causal features are key to achieve graph adversarial robustness, owing to their determinedness for labels and invariance across attacks.
To learn these causal features, we innovatively propose an \emph{\underline{I}nvariant causal \underline{DE}fense method against adversarial \underline{A}ttacks} (IDEA).
We derive node-based and structure-based invariance objectives from an information-theoretic perspective. 
%We design both node-based and structure-based invariance objectives based on the causalities to learn causal feature. 
%IDEA is provably a causally invariant defense across various attacks.
%IDEA offers a broad protection against attacks, ensuring strong predictability for labels and invariant predictability across attacks.
IDEA ensures strong predictability for labels and invariant predictability across attacks, which is provably a causally invariant defense across various attacks.  
Extensive experiments demonstrate that IDEA attains state-of-the-art defense performance under all five attacks on all five datasets.
The implementation of IDEA is available at \url{https://anonymous.4open.science/r/IDEA}.\end{abstract}


%\item We propose a novel graph adversarial immunization problem.

%$
%Y \Perp D \mid Z_{c}
%$

% Keywords

\maketitle

% Main text
%\section{}\label{}

% Use the style of numbering in square brackets.
% If nothing is used, default style will be taken.
%\begin{enumerate}[a)]
%\item 
%\item 
%\item 
%\end{enumerate}  

%\begin{itemize}
%\item 
%\item 
%\item 
%\end{itemize}  

%\begin{description}
%\item[]
%\item[] 
%\item[] 
%\end{description}  

% Figure
%\begin{figure}[<options>]
%	\centering
%		\includegraphics[<options>]{}
%	  \caption{}\label{fig1}
%\end{figure}

%\begin{table}[<options>]
%\caption{}\label{tbl1}
%\begin{tabular*}{\tblwidth}{@{}LL@{}}
%\toprule
%  &  \\ % Table header row
%\midrule
% & \\
% & \\
% & \\
% & \\
%\bottomrule
%\end{tabular*}
%\end{table}

% Uncomment and use as the case may be
%\begin{theorem} 
%\end{theorem}

% Uncomment and use as the case may be
%\begin{lemma} 
%\end{lemma}


\section{Introduction}
%Graph data are pervasive in the real world, such as citation networks~\cite{sen2008collective, hu2020open}, social networks~\cite{Zeng2019GraphSAINTGS}, financial networks~\cite{gai2010contagion}, and biological networks~\cite{hu2020open}. 
%In recent years, 
Graph neural networks (GNNs) have achieved immense success in numerous tasks and applications, including node classification~\cite{kipf2017semi}, recommendation~\cite{he_lightgcn_2020}, and fraud detection~\cite{Cheng2022Fraud}.
However, GNNs have been found to be vulnerable to adversarial attacks~\cite{zugner2018adversarial}, 
i.e., imperceptible perturbations on graph data can easily mislead GNNs into misprediction.
For example, in credit scoring, attackers add fake connections with high-credit customers to deceive GNNs~\cite{Jin2020AdversarialAA}, leading to loan fraud and severe economic losses. 
This vulnerability poses significant security risks,  hindering the deployment of GNNs in real-world scenarios. Therefore, defending against adversarial attacks is crucial and has attracted substantial research interests.

\begin{figure*}[t]
\centering
\captionsetup[subfigure]{font=scriptsize,labelfont=scriptsize}
\subcaptionbox{Node classification accuracy on clean and attacked graphs on Cora}
{\includegraphics[width = 0.63\textwidth]{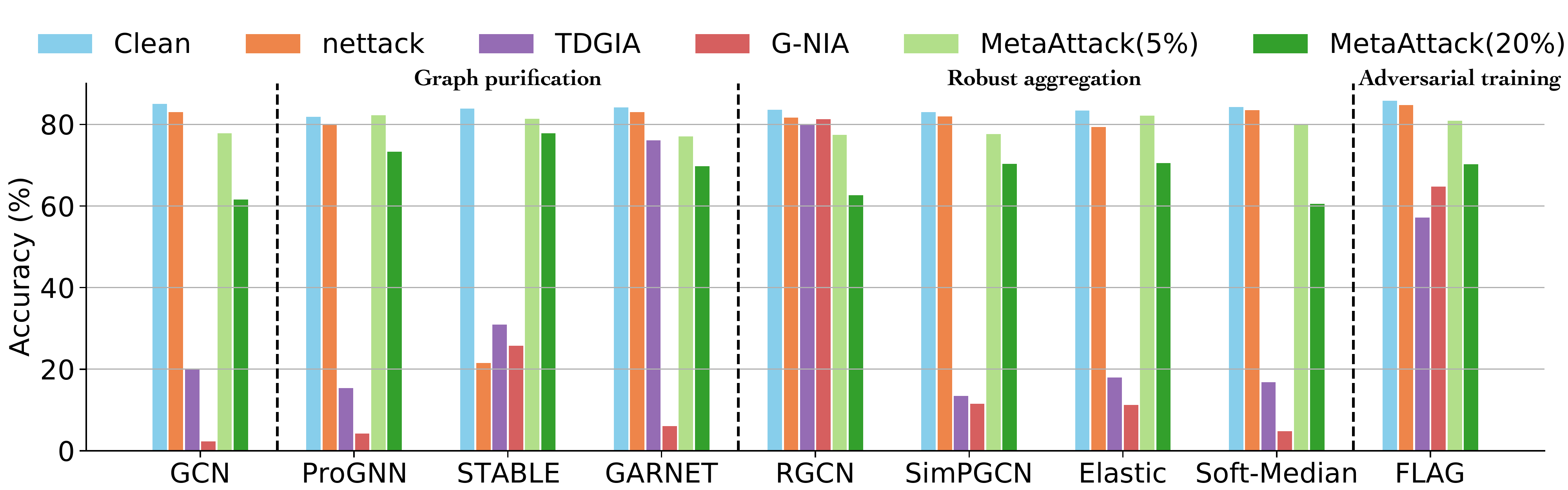}
} 
\hfill
\subcaptionbox{Clean: performance comparison}
{\includegraphics[width = 0.35\textwidth]{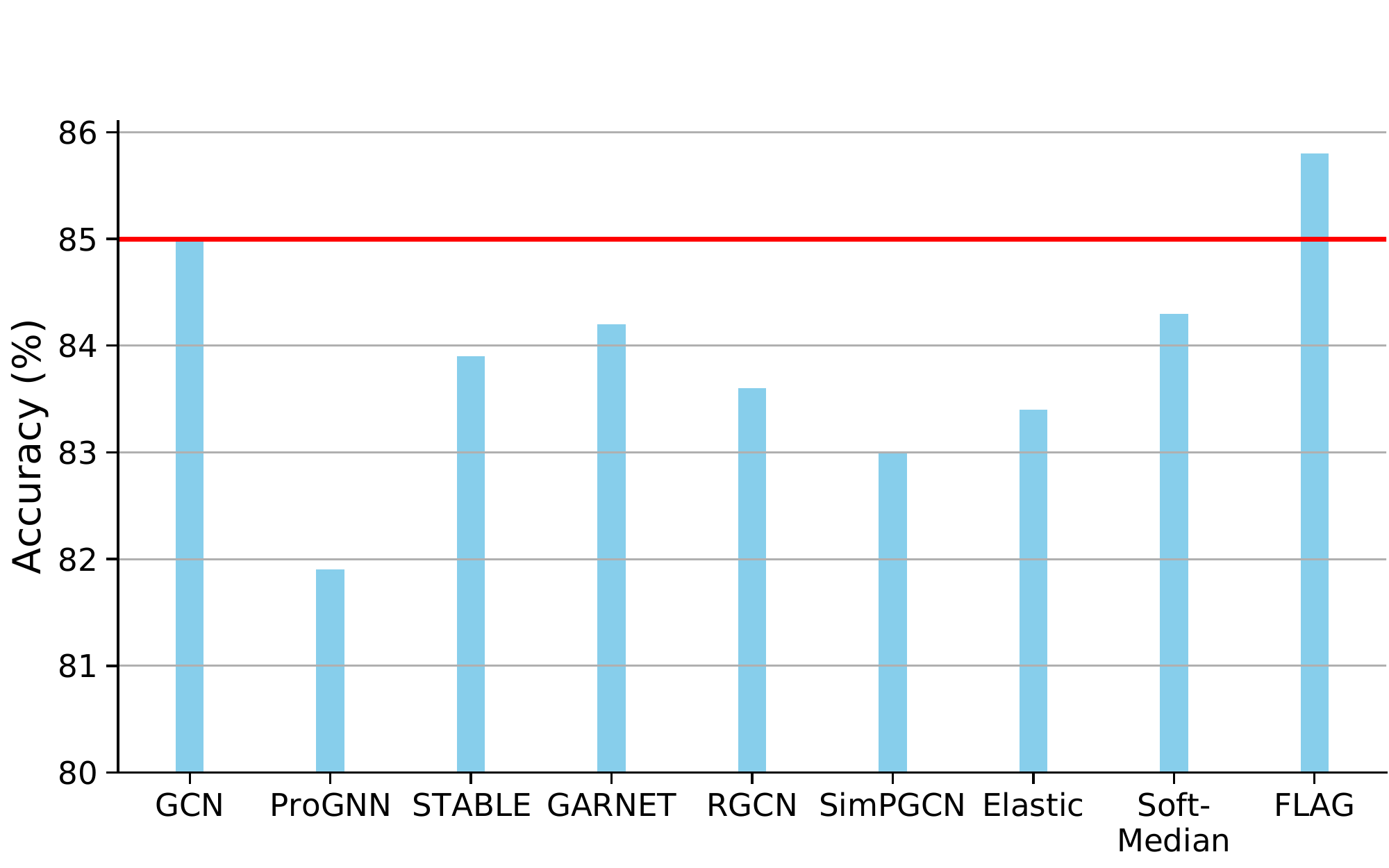}
} 
\caption{Limitation of existing methods: Defenses suffer performance degradation under various attacks and on clean graph.
% 5\% and 20\% denote perturbation rates of MetaAttack.
}
\label{fig:existing}
% \vspace{-6pt}
\end{figure*}

Existing defense methods, mainly including graph purification, robust aggregation, and adversarial training ~\cite{Jin2020AdversarialAA}, show effectiveness under specific attacks but lack broad protection across various attacks.
 Specifically, graph purification~\cite{Entezari2020AllYN} purifies adversarial perturbations by modifying graph structure, while robust aggregation~\cite{jin2021node} redesigns GNN structure to defend against attacks. Both methods rely on pre-defined heuristics such as local smoothness (e.g., SimPGCN~\cite{jin2021node}) or low rank (e.g., GARNET~\cite{GARNET22DengLF}). However, they are susceptible to unseen attacks, as shown in Figure~\ref{fig:existing} (a), ProGNN~\cite{Jin2020GraphSL} and SimPGCN fail against TDGIA~\cite{ZouTDGIA} since TDGIA deviates from the assumed heuristics.
The similar phenomenon is also observed in adversarial training, which tunes model on generated adversarial examples, resulting in vulnerability to unseen attacks. 
Moreover, modifying graph structure (ProGNN) or adding noise (RGCN~\cite{Zhu2019RobustGC}) even degrade performance on clean graphs, shown in Figure~\ref{fig:existing} (b).

% Moreover, modifying graph structure (STABLE~\cite{li2022reliable}) or adding noise (RGCN~\cite{Zhu2019RobustGC}) even degrade performance on clean graphs, in Figure~\ref{fig:existing} (b). Adversarial training (e.g., FLAG~\cite{kong2020flag}) tunes model on generated adversarial examples, which limits its scope to previously observed attacks.
%  Figure~\ref{fig:existing} (a) shows  adversarial training FLAG~\cite{kong2020flag} also exhibits unsatisfactory performance under TDGIA. 

% TODO : 缩略一下
To address the above limitations, we innovatively propose an invariant causal defense perspective. Note that while invariant causal methods have fueled a surge of research interests, they mainly focus on the independent samples~\cite{arjovsky2019invariant,LiCVPR}. However, these methods cannot be directly applied to solve adversarial robustness on graph due to its complex nature. 
On graph data, the interconnectedness (edges) among the samples (nodes) results in dependence among the nodes, which presents unique challenges.
In this paper, we design an interaction causal model~\cite{zhang2022causal} to capture causalities in graph adversarial attacks, tackling non-independent  nature of graph data. 
Our findings suggest that causal features are crucial for graph adversarial robustness due to their:
(1) \textbf{Strong predictability for labels}, since causal features determine labels; 
(2) \textbf{Invariant predictability across attacks}, as the causalities between causal features and labels remains consistent regardless of attacks.
These properties highlight causal features’ importance for robustness under unseen attacks.
%To learn causal features, we design an interaction causal model~\cite{zhang2022causal} to capture causalities between samples under graph attacks, addressing non-IID characteristics of graph data. 
% However, existing methods like structural causal models (SCMs) struggle with , specifically the interactions (e.g., edges) between samples (e.g., nodes).
%To overcome the challenge, we design an interaction causal model~\cite{zhang2022causal} to capture causalities between samples under graph attacks.  

Guided by our designed interaction causal model, we propose an \emph{\underline{I}nvariant causal \underline{DE}fense method against adversarial \underline{A}ttacks (IDEA)}.
By analyzing distinct characteristics of causal features, we derive node-based and structure-based invariance objectives considering both nodes and edges in graph.
Node-based invariance goal minimizes the conditional mutual information between label and attack given causal feature, based on the consistent causality between causal feature and label regardless of attacks.
While structure-based invariance goal is specially designed for graph structure, learning neighbors' causal feature to address the dependence among nodes.
IDEA is proved to be a causally invariant defense,  under the linear causal assumptions. 
It possesses strong and invariant predictability across attacks, offering broad protection.
Extensive experiments confirm its superiority, outperforming all baselines across five datasets and attack scenarios.
%Extensive experiments demonstrate that IDEA superiority significantly outperforms all baselines under both evasion and poisoning  attacks on five benchmark datasets.
%This emphasizes IDEA's strong and invariant predictability across attacks.

The main contributions of this paper are:
\begin{enumerate}[leftmargin=18pt]
	\item \emph{New perspective:} We propose an invariant causal defense perspective and design an interaction causal model to capture the causalities in graph adversarial attack, offering a new insights to the field.
	\item \emph{Novel methodology:} We propose IDEA to learn causal features for graph adversarial robustness. 
	We design two invariance objectives to learn causal features by modeling and analyzing the causalities in graph adversarial attacks. 
	\item \emph{Experimental evaluation:} Comprehensive experiments demonstrate that IDEA achieves state-of-the-art defense performance on all five datasets, highlighting its strong and invariant predictability.
\end{enumerate}

\section{Preliminary}
\label{sec:pre}
%We introduce the widely-used node classification task and graph neural networks. We also introduce the goal of graph adversarial attack and graph adversarial robustness.

%\subsection{GNN for Node Classification}
%\textbf{GNN for Node Classification.}
Given an attributed graph $G=(\mathcal{V}, \mathcal{E}, X)$, we denote $\mathcal{V} =\{1, 2, ..., n\}$ as node set, $\mathcal{E} \subseteq \mathcal{V} \times \mathcal{V}$ as edge set, and 
$X \in \mathbb{R}^{n \times d}$ as the attribute matrix with $d$-dimensional attributes. 
%We denote network structure as an adjacency matrix $D$. 
The class set $\mathcal{K}$ contains $K=|\mathcal{K}|$ classes.
The goal of node classification is to assign labels for nodes based on the node attributes and network structure by learning a GNN $f_{\theta}$~\cite{Jin2020AdversarialAA}.
The objective is:
$
	\min_{\theta} \sum_{i\in \mathcal{V}_\text{train}} [L(f_{\theta}(G)_i, Y_i)]
$, where $Y_i$ denotes the ground-truth label of node $i$.

%\textbf{Graph Neural Networks.}
%A typical GNN layer contains a message-passing function ($\textsc{Msg}$) and an updating operation ($\textsc{Upd}$)~\cite{li2022reliable}. Given a node $i$ and its neighborhood $\mathcal{N}_i$, the GNN initially applies the $\textsc{Msg}$ function to aggregate information from $\mathcal{N}_i$: $\bm{m}_i^l = \textsc{Msg}({\bm{h}_j^{l-1};j\in \mathcal{N}_i})$, where $\bm{m}_i^l$ is the  $\bm{h}_j^{l-1}$ denotes the hidden representation from the previous layer, and $\bm{h}_j^{0}=\bm{x}_k$. Then, the GNN updates the representation using the $\textsc{Upd}$ function: $\bm{h}_i^l = \textsc{Upd}(\bm{m}_i^l, \bm{h}_i^{l-1})$, which is usually a sum operation. 

%The final node representation for $i$ is $h_i^L$, i.e., the output of the $L$-th layer.

% \begin{figure}[htbp]
% \centering
% 	\subcaptionbox{SCM: single node}
% 	{\includegraphics[width = 0.08\textwidth]{iscm_one}
% 	}
% 	\hfill
% 	\subcaptionbox{Interaction SCM: graph data with interconnection}
% 	{\includegraphics[width = 0.17\textwidth]{iscm_two}
% 	}
% 	\hfill
% 	\subcaptionbox{Interaction SCM: graph adversarial attack}
% 	{\includegraphics[width = 0.17\textwidth]{iscm_atk}
% 	} 
% \caption{The structural causal models of the generation process of graph data and graph adversarial attack.}
% \label{fig:scm}
% \end{figure}

%\textbf{Graph adversarial attacks.}
The graph adversarial attack aims to find a perturbed graph $\hat{G}$ that maximizes the loss of GNN model~\cite{Jin2020AdversarialAA}: 
%on the target nodes $\mathcal{V}$:
\begin{equation}
	\begin{aligned}
&\max_{\hat{G} \in \mathcal{B}(G)} \sum_{i\in \mathcal{V}} [L(f_{\theta^*}(\hat{G})_i, Y_i)] \\
s.t., \quad &\theta^* = \arg \min_{\theta} \sum_{i\in \mathcal{V}} [L(f_{\theta}(G_{\text{train}})_i, Y_i)].
\end{aligned}
\end{equation}
Here, perturbed graph $\hat{G}$ is chosen from the admissible set $\mathcal{B}(G)$, where the perturbed nodes, edges, and node attributes should not exceed the budget~\cite{Jin2020AdversarialAA}.
%$\mathcal{B}(G,\Delta) = \left\{\hat{G}=(\hat{V},\hat{E},\hat{X})|\|\hat{V}-V\|_{0}\leq \Delta_V,  \|\hat{X}-X\|_{0} \leq \Delta_X \right\}.$
$G_{\text{train}}=G$ in evasion attacks, and $G_{\text{train}}=\hat{G}$ in poisoning attacks. 

%\textbf{Graph adversarial robustness.}
Defense methods aim to improve graph adversarial robustness.
The goal can be formulated as:
\begin{equation}
	\min_{\theta} \max_{\hat{G} \in \mathcal{B}(G)} \sum_{i\in \mathcal{V}} [L(f_{\theta}(\hat{G})_i, Y_i)].
	\label{eq:problem}
\end{equation}
%The goal is to minimize upper-bound of ERM risk.
%\textbf{Limitations.}
%However, it is challenging to achieve graph adversarial robustness against any attack in the constraints.
Existing defense methods suffer performance degradation under various attacks or on clean graphs. Adversarial training~\cite{kong2020flag}  
% can only be optimized for the training adversarial examples, limiting their ability to 
generalizes poorly to unseen adversarial attacks.  
While graph purification and robust aggregation are designed based on specific heuristic priors, such as local smoothness~\cite{li2022reliable,jin2021node} and low rank~\cite{Jin2020GraphSL,Entezari2020AllYN}. They are only effective when attacks satisfy these priors.
% and can even damage performance on clean graphs.
Hence, there is an urgent need to design a defense method that performs well both on clean graphs and across various attacks.
%Existing methods have limitations in achieving graph adversarial robustness.
%Adversarial training methods can only be optimized for the maximum among the seen adversarial examples. But in essence, is still an interpolation~\cite{arjovsky2019invariant} and cannot generalize to the unseen adversarial examples.
%Robust GNNs or graph purification methods design models according to the heuristic prior on certain adversarial examples, so that only effective to the adversarial examples that satisfy this prior.
%Hence, it urges us to design a defense method to achieve graph adversarial robustness against various attacks. 
%So far, defending against attacks from a causal invariant perspective is left unexplored so far.

\section{Methodology}
We first model the causalities between causal features and other variables in graph adversarial attacks.
Based on this causal analysis, we propose an \underline{I}nvariant causal \underline{DE}fense method against \underline{A}ttacks (IDEA) to learn causal features. 
% We design two invariance goals for graph adversarial robustness. 

\begin{figure*}
% \vspace{-5pt}
    \centering
    \includegraphics[width = 0.95\textwidth]{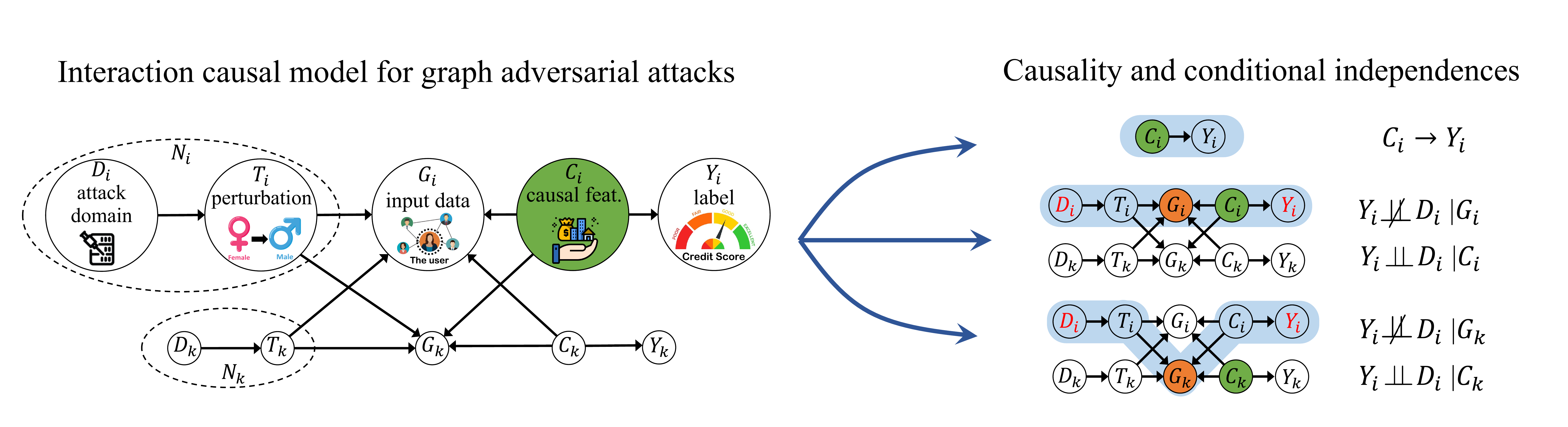}
    \caption{\textbf{Left}: {Interaction causal model for graph adversarial attacks.} \textbf{Right}: Causality and conditional independences.}
    \label{fig:scm}
    % \vspace{-10pt}
\end{figure*}

%%\begin{wrapfigure}{r}{0.4\textwidth}
%\begin{figure}[t]
%%% \vspace{-4mm}
%\centering
%\captionsetup[subfigure]{font=scriptsize,labelfont=scriptsize}
%\subcaptionbox{}
%{\includegraphics[width = 0.28\textwidth]{scm_graphG.jpg}
%\label{subfig:scm_graph}
%} 
%\subcaptionbox{}
%{\includegraphics[width = 0.22\textwidth]{scm_atkG.jpg}
%\label{subfig:scm_atk}
%}
%\subcaptionbox{}
%{\includegraphics[width = 0.22\textwidth]{scm_path1G.jpg}
%\label{subfig:scm_path}
%}
%\subcaptionbox{}
%{\includegraphics[width = 0.22\textwidth]{scm_path2G.jpg}
%\label{subfig:scm_path}
%}
%%% \vspace{-3pt}
%\caption{(a) Interaction causal model of graph data.
%%, where the variables related to $i$ take the fraud detection icon as an example.
%(b) Interaction causal model of graphs adversarial attack.
%%, where the perturbation $T_i$ influences the non-causal features, is determined by attack domain $D_i$.
%(c,d) The causal paths between $Y_{i}$ and  $D_{i}$ conditioned on $G_i$ (c) and $G_k$ (d).
%}
%\label{fig:scm}
%%% \vspace{-10pt}
%\end{figure}
%%\end{wrapfigure}

\subsection{Interaction Causal Model}
\label{sec:causal}

To model the dependence characteristics in graph data, namely the interactions (e.g. edges) between samples (e.g. nodes), we design an interaction causal model with explicit variables~\footnote{Note that the explicit variable $X_i$ refers to the event of one specific sample $i$, and the generic variable $X$ is the event of all samples.} to capture the causality between different samples~\cite{zhang2022causal} under graph adversarial attacks.

% %Before diving into details, we first consider the causal mechanism in graph data as a preparation.

Figure~\ref{fig:scm} (left) illustrate an example involving two connected nodes $i$ and $k$. We inspect the causal relationships among variables: input data $G_i$ (node $i$'s ego-network), label $Y_i$, causal feature $C_i$, perturbation $T_i$, attack domain $D_i$, and those variables of neighbor node $k$. 
%We introduce two latent features $C_i$ and $N_i$ as abstractions that determine the observed $G_i$ and $Y_i$, which have been similarly assumed in invariant learning works~\cite{li2022invariant,ren2022dice}. 

We introduce the latent causal feature $C_i$ as an abstraction that causes both input ego-network $G_i$ and label $Y_i$.
For example, in credit scoring, $C_i$ represents the financial situation, which determines both $G_i$ (including personal attributes and friendships)  and credit score $Y_i$. 
%While the input $G_i$ is not only determined by $C_i$, but also by other features, which we denote as the non-causal feature $N_i$ (e.g., the number of friends). 
Besides, the causal feature $C_i$ influences neighbor $G_k$ due to network structure, aligning with GNN studies~\cite{kipf2017semi,velickovic2018graph}.
%Next, we devote to model the causality in graph adversarial attacks.
We model graph adversarial attack with perturbation $T_i$ and attack domain $D_i$ which is a latent factor that determines $T_i$, as shown in Figure~\ref{fig:scm} (left).
Attack domain $D_i$ denotes attack categories based on characteristics, like attack type or attack strength.
Here, $D_i$ and $T_i$ are considered as non-causal features $N_i$ associated to attack, and we strive to exclude their influence. 
Perturbation $T_i$ may impact the neighbor ego-network $G_k$  due to edges between nodes.

%The concept of the attack domain is introduced primarily to learn the causal invariant feature that remains invariant predictability across different attack domains.

We analyze these causalities and find that: (i) Causal feature $C$ determines label $Y$, indicating causal feature's \textbf{strong predictability} for label; (ii) The $C-Y$ causality remains unchanged across attack domains, indicating the causal feature maintains \textbf{invariant predictability} across attack domains.
These properties make causal features beneficial in enhancing graph adversarial robustness.
Specifically, strong predictability enables good performance on clean graphs, while invariant predictability maintains performance under attacks.
Meanwhile, the impact of attacks including $D_i$ and $T_i$ should be eliminated.
Based on the above intuition, we aim to design a method to learn causal feature $C$ and reduce the influence of non-causal feature to defend against attacks.

\subsection{IDEA: Invariant causal DEfense method against adversarial Attack}
\label{sec:idea}
We  propose IDEA
% \emph{\underline{I}nvariant causal \underline{DE}fense against adversarial \underline{A}ttack (IDEA)} 
to learn causal features. 
% Specifically, we design invariance objectives based on the unique characteristics of causal features and derive approximate losses. Meanwhile, to reveal the diverse influence of attack domains, we introduce a domain learner for suitable domain partitioning. The overall framework of IDEA is also provided.
% To achieve this, we design invariance objectives based on the distinctive properties of causal features and derive approximate losses. Additionally, we introduce a domain learner to effectively partition domains, accounting for the varying impact of attack domains. A comprehensive framework of IDEA is also presented.
Our approach involves designing invariance objectives based on the distinctive properties of causal features and approximating losses accordingly. 
%To reveal the diverse influence of attack domains, we introduce a domain learner for partitioning. 
% We present the overall framework of IDEA.

%According to our theoretical analysis, IDEA optimizes adversarial domain partition to construct diverse domains to realize robustness against all adversarial attacks.
%Theoretical proof and extensive experiments demonstrate that our method can defend against different attacks and achieve graph adversarial robustness.

\subsubsection{Invariance Objective}

To learn causal features, we design invariance objectives by analyzing causality and conditional independences with $d$-separation~\cite{pearl10causalinfer} in Figure~\ref{fig:scm}.
The observations are:
\begin{enumerate}[leftmargin=12pt]
	\item $C_i \rightarrow Y_i$:~\label{obs:1} Causal feature $C$ determines label $Y$ (first path in Figure~\ref{fig:scm} (right)).
	\item  $Y_i \not\! \indep D_i \mid G_i$:~\label{obs:2} $Y$ and attack domain $D$ are associated given $G$ (second path in Figure~\ref{fig:scm} (right)), since $G$ is a collider\footnote{A collider is causally influenced by two variables and blocks the association between the variables that influence it. Conditioning on a collider opens the path between its causal parents~\cite{pearl10causalinfer,pearl2009causal}. In our case, $C_i$ and $T_i$ are associated conditioned on $G_i$, making $Y_i$ and $D_i$ associated, conditioned on $G_i$.} between $Y$ and $D$. 
%	\item This means $G_i$ is influenced by attacks.
	\item $Y_i \indep D_i \mid C_i$:~\label{obs:3} The $C-Y$ causality remains unchanged across attack domain $D$, i.e., $C$ has invariant predictability for $Y$ across various attack domains.%	\item  of $Y_{i}$ and  $D_{i}$ conditioning on $N_{i}$ is .
	\item  $Y_{i} \not\! \indep D_{i} \mid G_{k}$:~\label{obs:4} $Y_{i}$ and  $D_{i}$ are associated given the neighbor's ego-network $G_{k}$ (third path in Figure~\ref{fig:scm}(right)), since $G_k$ is a collider of $Y_i$ and $D_i$. It is crucial to recognize that this association is unique and arises from the dependencies between nodes which are inherent to the graph structure.
	%	\item where $G_{k}$ is also influenced by attacks. 
	\item $Y_{i} \indep D_{i} \mid C_{k}$:~\label{obs:5} $Y_{i}$ and $D_i$ are independent conditional on neighbor's causal feature $C_{k}$.
%	that label $Y_{i}$ is independent of the attack domain $D_{i}$ conditioned on the causal feature $C_k$ of neighbor $k$. 
\end{enumerate}

%We hope $\Phi(G)$ to learn causal feature $C$.
Based on these observations, we analyze the characteristics of $C$ and propose three goals from the perspective of mutual information $I$ to learn causal feature $C$. Let $\Phi$ represent the feature encoder.
\begin{itemize}[leftmargin=9pt]
%	\item According to (1,2), $A \indep Z$ means we should learn the shared feature $Z$ across adversarial attacks.
	\item  \textbf{Predictive goal}: $\max_{\Phi} I\left(\Phi(G), Y\right)$ to guide {$\Phi(G)$} to have strong predictability for $Y$ in (\ref{obs:1}).
	\item  \textbf{Node-based Invariance goal}: $\min_{\Phi}  I\left(Y, D \mid \Phi(G)\right)$. By comparing the conditional independences regarding $G_i$ in (\ref{obs:2}) and $C_i$ in (\ref{obs:3}), we propose this goal to guide {$\Phi(G)$} learning $C$ and excluding the influence of attack, to obtain invariant predictability across different attack domains. %	\item based on (2,3). Intuitively, $C$ in order to expose the various effects of attacks.
	\item \textbf{Structure-based Invariance goal}: ${\small\min_{\Phi}  I(Y, D \mid \Phi(G)_\mathcal{N})}$ Through comparing $G_k$ in (\ref{obs:4}) and $C_k$ in (\ref{obs:5}), we propose this goal to guide {$\Phi(G)_{\mathcal{N}}$} learning the causal feature for neighbor {$\mathcal{N}$}. This goal is specifically designed for the graph data, taking into account intrinsic the dependence among nodes. It leverages the graph structure to better learn causal features. 
\end{itemize}
% Note that the attack domain $D$ should be diverse to expose the various influence of attacks to learn the invariant causal feature, as detailed Section~\ref{sec:domain}
To sum up, the objective can be formulated as:
\begin{equation}
%\small
\begin{aligned}
\max _{\Phi}\quad & I\left(\Phi(\hat{G}^{*}), Y\right)-
\left[I\left(Y, D \mid \Phi(\hat{G}^{*})\right)+ I\left(Y, D \mid \Phi(\hat{G}^{*})_{\mathcal{N}}\right)\right] \\ 
\text { s.t. } \quad & \hat{G}^{*}=\arg\min_{\hat{G} \in \mathcal{B}(G)} I\left(\Phi(\hat{G}), Y\right), 
% \quad D=\arg\min \sum _{D_1 \neq D_2} \text{Sim} \left( r(Z^{D_1}), r(Z^{D_2}) \right)
\label{eq:goal}
\end{aligned}
\end{equation}
%Here, $\hat{G}$ is the adversarial example, $D$ is the attack domain. 
Here, attack domain $D$ is used to expose the difference of attack influence on feature, 
mitigate
thereby learning the causal features that are invariant across attacks. 
The objective guides IDEA to learn causal feature with strong predictability and invariant predictability across attack domains, as well as exclude the impact of attacks.
The capability of this objective relies on the diversity of attack domain $D$ (Section~\ref{sec:domain}). %Proposition~\ref{theo:achiev} demonstrates the effectiveness of our objective, assuming linear causalities.
However, two challenges persist in solving Eq.~\ref{eq:goal}: i) The objective is not directly optimizable since estimating mutual information of high-dimensional variables is difficult. 
ii) It is unknown that how to design attack domain $D$. Intuitively, diverse $D$ can expose the difference of attack influence on features and promote learning inviarant causal feature.

To address the challenge i), Section \ref{sec:loss} presents the loss approximations for our proposed objectives.
To tackle the challenge ii), Section \ref{sec:domain} provides domain construction to learn the attack domain.
%, which are required to expose influence of the attack and promote learning invariant causal features. 
% Constructing differentiated domains presents additional problem to address.

%However, there are still two obstacles to solving this objective function: 
%i) The objective cannot be optimized directly, since it is difficult to estimate the mutual information of high-dimensional variables.
%ii) There is a lack of diverse attack domains which can help to exclude non-causal features 
% affected by the attack, facilitating better learning of causal feature. How to construct differentiated domains and generate adversarial examples is a problem.
%%There is no information about which attack domain a sample (i.e., a node) belongs to In the field of graph adversarial attacks, intuitive

\begin{figure*}[t]
\centering
% \vspace{-5pt}
\includegraphics[width = 0.9\textwidth]{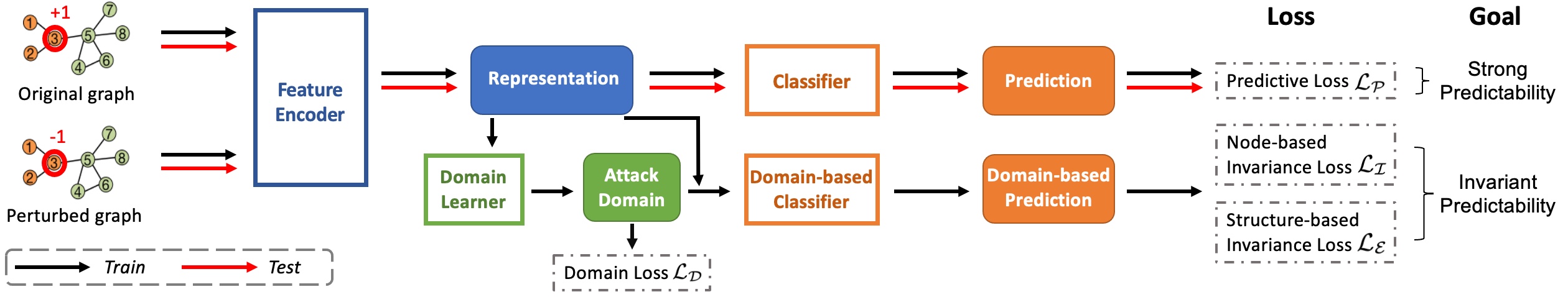}
\caption{Overall architecture of our IDEA method. {IDEA contains feature encoder, classifier, domain-based classifier, and domain learner.} The black arrows denote the workflow of IDEA during training, while the red arrows denote how IDEA predicts in the test phase. }
\label{fig:frame}
% \vspace{-5pt}
\end{figure*}

\subsubsection{Loss Approximation}
\label{sec:loss}
% We develop tractable losses for the above goals. 
%and let $Z=\Phi(\hat{G}^*)$ for simplicity.
%$z$ 
%, and $x=i \in \mathcal{V}$  

Let {\small$Z=\Phi(\hat{G}^{*})$} denote the representation to learn $C$. 

\paragraph{Predictive loss.}

The predictive goal is {\small $I(Z,Y) =\mathbb{E}_{p(y,z)} \log \frac{p(y|z)}{p(y)} $}. 
Unfortunately, it is challenging to directly compute the distribution $p(y|z)$. To overcome this, we introduce $q(y|z)$ as a variational approximation of $p(y|z)$. Similar to~\cite{deepvib2017,li2022invariant}, we derive a lower bound of predictive goal: $\mathbb{E}_{p(x,y)} \mathbb{E}_{p(z|x)} \log q\left(y \mid z\right)$, 
%\begin{equation}
%\small
%\begin{aligned}
%I(Z,Y) =\mathbb{E}_{p(y,z)} \log \frac{p(y|z)}{p(y)}  
%\geq 
%\mathbb{E}_{p(x)} \mathbb{E}_{p(y|x)} \mathbb{E}_{p(z|x)} \log q\left(y \mid z\right) \approx
%\mathbb{E}_{p(x,y)}\mathbb{E}_{p(z|x)} \log q\left(y \mid z\right),
%\label{eq:lower} 
%\end{aligned}
%\end{equation}
where $x$ is the input ego-network. Consequently, we can maximize $I(Z,Y)$ by maximizing the lower bound.

The lower bound involves two distributions, $p(z|x)$ and $q(y|z)$ that need to be solved. For $p(z|x)$, we employ neural network $h$ as a realization of $\Phi$ to learn representation. We assume a Gaussian distribution $p(z \mid x)=\mathcal{N}\left(z \mid h^{\mu}(x), h^{\Sigma}(x)\right)$, where  $h^{\mu}$ and $h^{\Sigma}$ output the mean and covariance matrix, respectively~\cite{li2022invariant,deepvib2017}.  
Subsequently, we leverage a re-parameterization technique~\cite{reparmet2014} to tackle non-differentiability:  $z=h^{\mu}(x)+\epsilon  h^{\Sigma}(x)$, where $\epsilon$ is a standard Gaussian random variable. In general, encoder $h$ contains GNN and re-parameterization, outputting $z=h(x, \epsilon)$. For $q(y|z)$, we use a neural network $g$ as classifier to learn the variation distribution $q(y|z)$. With $p(z|x)$ and $q(y|z)$, we obtain the predictive loss $\mathcal{L}_\mathcal{P}$:
\begin{equation}
%\small
%	\min_{g,h} \mathcal{L}_\mathcal{P}\left(g(h(\hat{G}^*))\right) =
	\min_{g,h} \mathcal{L}_\mathcal{P}\left(g, h, \hat{G}^*\right) =
	 \min_{g,h} \sum_{i\in \mathcal{V}_l} L(g(h(\hat{G}^*)_i), Y_i).
\end{equation}
%where $h$ is a GNN encoder (realization of $\Phi$) to learn causal feature $C$, and $g$ is an NN classifier. 

\paragraph{Node-based invariance loss.}

The node-based invariance goal strategically minimizes the conditional mutual information between the label and attacks, conditional on the causal features. 
This is rooted in the principle of consistent causality, which posits that the relationship between causal features and labels remains stable irrespective of adversarial attacks. 

%Emphasizing the advantages, this goal fortifies the model against adversarial manipulation by anchoring the predictive process in the steadfast causality between features and labels. It effectively disentangles the influence of attacks from the predictive mechanism, thereby enhancing the model's resilience and reliability. The conditional mutual information, which serves as a quantifiable measure for this goal, is specified as follows:

The conditional mutual information is defined as:
\begin{equation*}
%\small
	\begin{aligned}
I(Y,D|Z)=&\ \mathbb{E}_{p(z)}\left[ \mathbb{E}_{p(y, d\mid z)} \left[\log \frac{p(y, d \mid z)}{p(d \mid z) p(y \mid z)}\right]\right]\\
=&\ \mathbb{E}_{p(z)}\left[ \mathbb{E}_{p(y, d \mid z)}[\log p(y \mid z, d)-\log p(y \mid z)]\right].
\end{aligned}
\end{equation*}
To approximate $p(y \mid z, d),\ p(y \mid z)$, we also employ two variational distributions $q_d(y \mid z, d)$ and $q(y \mid z)$. This allows us to obtain an estimation of $I(Y,D|Z)$:
\begin{equation*}
%\small
\hat{I}(Y,D|Z)= \mathbb{E}_{p(z)}\left[ \mathbb{E}_{p(y, d \mid z)}[\log q_d(y \mid z, d)-\log q(y \mid z)]\right].
\end{equation*}

Similar to CLUB~\cite{club2020,LiCVPR}, we minimize $\mathbb{E}_{p(z,d)} KL[p(y \mid z, d) \| q_{d}(y \mid z, d)]$ to make $\hat{I}(Y,D|Z)$ as an upper bound on $I(Y,D|Z)$.
We prove that  minimizing both $\mathbb{E}_{p(z,d)} KL[p(y \mid z, d) \| q_{d}(y \mid z, d)]$ and  $\hat{I}(Y,D|Z)$   minimizes our goal $I(Y,D|Z)$.
\begin{proposition}
\label{prop:cond_info}
	The node-based invariance goal $I(Y,D|Z)$ reaches its minimum value if $\mathbb{E}_{p(z,d)} KL[p(y \mid z, d) \| q_{d}(y \mid z, d)]$ and $\hat{I}(Y,D|Z)$ are minimized
\end{proposition}
The proof is in Appendix~\ref{proof:cond_info}. 
%Then we use a neural network $g_d$ as classifier to learn $q_d(y \mid z, d)$. We optimize the KL divergence $\mathbb{E}_{p(z,d)} KL[p(y \mid z, d) \| q_{d}(y \mid z, d)]$ by minimizing $\sum_{i\in \mathcal{V}} L\left(g_d(h(\hat{G}^*)_i,D_i), Y_i\right)$, and optimize $\hat{I}(Y,D|Z)$ by minimizing $\sum_{i\in \mathcal{V}}\left[ L\left(g(h(\hat{G}^*)_i), Y_i) - L(g_d(h(\hat{G}^*)_i,D_i), Y_i\right)\right]$. 
We utilize a neural network $g_d$ to learn $q_d(y \mid z, d)$, optimizing the KL divergence $\mathbb{E}_{p(z,d)} KL[p(y \mid z, d) \| q_{d}(y \mid z, d)]$ through minimizing the loss $\sum_{i\in \mathcal{V}} L(g_d(h(\hat{G}^*)_i,D_i), Y_i)$. Additionally, we refine $\hat{I}(Y,D|Z)$ by reducing $\sum_{i\in \mathcal{V}}[ L(g(h(\hat{G}^*)_i), Y_i) - L(g_d(h(\hat{G}^*)_i,D_i), Y_i)]$. 
The node-based invariance loss $\mathcal{L}_\mathcal{I}$:
\begin{equation}
%\small
\begin{aligned}
%	\min_{g,g_d,h}& \mathcal{L}_\mathcal{I}\Bigl(g\bigl(h(\hat{G}^*)\bigr), g_d\bigl(h(\hat{G}^*),D\bigr)\Bigr)\\
	&\min_{g,g_d,h} \mathcal{L}_{\mathcal{I}}\left(g, g_d, h, \hat{G}^*, D\right)\\
	= 
	&\min_{g,g_d,h} \sum_{i\in \mathcal{V}}L\Bigr(g_d\bigl(h(\hat{G}^*)_i,D_i\bigr), Y_i\Bigr)
	+  \alpha \left[L\Bigr(g\bigl(h(\hat{G}^*)_i\bigr), Y_i\Bigr) - L\Bigr(g_d\bigl(h(\hat{G}^*)_i,D_i\bigr), Y_i\Bigr)\right],
\end{aligned}
\end{equation}
where coefficient $\alpha$ balances the two terms.
%where coefficient $\alpha$ is a hyper-parameter to balance the two terms. 

\paragraph{Structure-based invariance loss.}
% Intuition

To address the inherent dependencies between nodes in graph data, we design the structure-based invariance goal. 
This objective is crafted with the unique topology of graph data in mind, specifically considering its dependence nature among nodes. 
The intuition behind this goal is to harness the relational information among nodes to learn causal features of neighborhood $\mathcal{N}$. By doing so, we can capture the essence of the graph structure, which is pivotal for robust and generalizable model performance.

To operationalize this goal, we optimize the structure-based invariance loss, $\mathcal{L}_\mathcal{E}$, which serves as a guiding metric for the learning process:
\begin{equation}
%\small
\begin{aligned}
%	\min_{g,g_d,h}& \mathcal{L}_\mathcal{E}\Bigl(g\bigl(h(\hat{G}^*)_{\mathcal{N}}\bigr), g_d\bigl(h(\hat{G}^*)_{\mathcal{N}},D\bigr)\Bigr)\\
	&\min_{g,g_d,h}
% 	\mathcal{L}_\mathcal{E}\left(g\circ h, g_d\circ h, D\right)\\
	\mathcal{L}_\mathcal{E}\left(g, g_d, h, \hat{G}^{*} D\right)\\
	= 
	&\min_{g,g_d,h} \sum_{i\in \mathcal{V}, k \sim \mathcal{N}_i}L\Bigr(g_d\bigl(h(\hat{G}^*)_k,D_i\bigr), Y_i\Bigr) 
	+  \alpha \left[L\Bigr(g\bigl(h(\hat{G}^*)_k\bigr), Y_i\Bigr) - L\Bigr(g_d\bigl(h(\hat{G}^*)_k,D_i\bigr), Y_i\Bigr)\right],
\end{aligned}
\end{equation}
where $k$ is the sampled node from the neighbors $\mathcal{N}_i$ of node $i$.
This loss function is designed to minimize any extraneous variability that does not contribute to the true signal of the data, thereby reinforcing the model's predictive power and stability in the face of changing graph dynamics.

\paragraph{Overall loss.}

In summary, the overall loss function consists of predictive loss and two newly proposed invariance losses:
%\begin{equation}
%\small
%\begin{aligned}
%\min_{g,g_d,h}& \mathcal{L}_\mathcal{P}\Bigl(g\bigl(h(\hat{G}^*)\bigr)\Bigr) 
%+ \alpha \Bigl[\mathcal{L}_\mathcal{I}\Bigl(g\bigl(h(\hat{G}^*)\bigr), g_d\bigl(h(\hat{G}^*),D\bigr)\Bigr) 
%+ \mathcal{L}_\mathcal{E}\Bigl(g\bigl(h(\hat{G}^*)_{\mathcal{N}}\bigr), g_d\bigl(h(\hat{G}^*)_{\mathcal{N}} ,D\bigr)\Bigr)\Bigr] \\
%\text { s.t. } \ & \hat{G}^{*}=\arg\max_{\hat{G}}\mathcal{L}_\mathcal{P}\Bigl(g\bigl(h(\hat{G}^*)\bigr)\Bigr).
%\end{aligned}
%\end{equation}
%
\begin{equation*}
%\small
\begin{aligned}
\!\min_{g,g_d,h}& \mathcal{L}_\mathcal{P}\left(g, h, \hat{G}^*\right) 
+\mathcal{L}_\mathcal{I}\left(g, g_d, h, \hat{G}^*, D\right)
% \right. \\ &\left. 
+ \mathcal{L}_\mathcal{E}\left(g, g_d, h, \hat{G}^*, D\right)\\
\text { s.t. } 
\ & \hat{G}^{*}=\arg\max_{\hat{G} \in \mathcal{B}(G)} \mathcal{L}_\mathcal{P}\left(g, h, \hat{G}\right).
\end{aligned}
\label{eq:loss}
\end{equation*}

\subsubsection{Domain Construction}
 \label{sec:domain}
How to design attack domain $D$ remains critical. Straightforward ways such as categorizing by attack type or strength can yield very few, non-diverse domains.
Intuitively, attack domains should be both sufficiently numerous~\cite{rosenfeld2020risks,arjovsky2019invariant} and distinct from each other~\cite{creager2021environment,arjovsky2019invariant} to reveal the various effects of attacks. 
% The following critical question arises: how can we construct attack domains $D$ that enable the model to learn causal features which remain invariant predictability across attack domains? 
% Our aim is to learn the representation with invariant predictability across domains, corresponding to the desired causal feature. 
To this end, we leverage a neural network $s$ as domain learner to learn attack domain. 
Here, $s$ allows for the adjustable number of domains. We ensure the domain diversity by minimizing the co-linearity between the samples from different domains.
%For the requirement of the dimension of spanning space, the training domains should be diverse, which means that the co-linearity between the training domains is as small as possible.
We adopt Pearson correlation coefficient (\textbf{PCCs}) to measure of linear correlation between two sets of data.
The loss function $\mathcal{L}_\mathcal{D}$: 
\begin{equation}
%\small
\begin{aligned}
	 \min_{s}  \mathcal{L}_\mathcal{D}\left( s, h \right)
	 = &\min_{s} \sum_{D, D^{\prime}\in \mathcal{D}, D\neq D^{\prime}} \textbf{PCCs}\left(r^D, \rho(\hat{G})^{D^{\prime}}\right)\\
% & m^a=\mathbb{E}_{i\in \mathcal{V}^{D}}\left[h(\hat{G}^*)_i \left(g(h(\hat{G}^*)_i)-Y_i\right)\right]-\mathbb{E}_{i\in \mathcal{V}^{D}, \epsilon}\left[ h(\hat{G}^*)_i\epsilon\right], \\
 r^D=&\mathbb{E}_{i\in \mathcal{V}^{D}}\left[h(\hat{G}^*)_i \left(g(h(\hat{G}^*)_i)-Y_i\right)\right], \\
 \mathcal{V}^{D} =&\left\{i |(s(h(\hat{G}^*)_i)=D\right\},
 \label{eq:ld}
\end{aligned}
\end{equation}
where attack domain $D$ is in the form of one-hot vector to categorize adversarial samples, $\mathcal{D}$ is the attack domain set, $\mathcal{V}^{D}$ denotes nodes assigned to domain $D$ by learner $s$, $r^D$ denotes the representation of $\mathcal{V}^{D}$. 
%Intuitively, $\left(h(\hat{G}^*)_i g(h(\hat{G}^*)_i)-Y_i\right)$ implies assigning larger weight to the misclassified nodes (i.e.,$g(h(\hat{G}^*)_i)-Y_i\neq0$).
The form of $r^D$ aids in proving IDEA achieving adversarial robustness (Proposition~\ref{theo:achiev}).

% We implement the domain learner $s$ using neural networks. 
\subsubsection{Overall Framework}

According to the above analysis, the overall loss function of IDEA is formulated as:
%\begin{equation}
%\small
%\begin{aligned}
%\!\min_{g,g_d,h}& \mathcal{L}_\mathcal{P}\left(g(h(\hat{G}^*))\right) 
%+\alpha\left[ \mathcal{L}_\mathcal{I}\left(g(h(\hat{G}^*)), g_d(h(\hat{G}^*),s^*(h(\hat{G}^*)))\right)
% \right. \\ &\left. 
%+ \mathcal{L}_\mathcal{E}\left(g(h(\hat{G}^*)), g_d(h(\hat{G}^*)_{\mathcal{N}}, s^*(h(\hat{G}^*)))\right)\right]\\
%\text { s.t. } 
%\ & \hat{G}^{*}=\arg\max_{\hat{G}} \mathcal{L}_\mathcal{P}\left(g(h(\hat{G}^*))\right),
%\quad s^{*}=\arg\min_{s} \mathcal{L}_\mathcal{D}\left( s \right).
%\end{aligned}
%\label{eq:final_loss}
%\end{equation}
%
\begin{equation}
%\small
\begin{aligned}
\!&\min_{g,g_d,h} \mathcal{L}_\mathcal{P}\left(g, h, \hat{G}^*\right) 
+\mathcal{L}_\mathcal{I}\left(g, g_d, h, \hat{G}^*, s^*\right)
% \right. \\ &\left. 
+ \mathcal{L}_\mathcal{E}\left(g, g_d, h, \hat{G}^*, s^*\right)\\
\!&\text { s.t. } 
  \hat{G}^{*}=\arg\max_{\hat{G} \in \mathcal{B}(G)} \mathcal{L}_\mathcal{P}\left(g, h, \hat{G}\right),
\ s^{*}=\arg\min_{s} \mathcal{L}_\mathcal{D}\left( s, h \right).
\end{aligned}
\label{eq:final_loss}
\end{equation}
The overall architecture of IDEA is illustrated in Figure~\ref{fig:frame}.
The IDEA model consists of four parts: an encoder $h$  to learn the node representation, i.e., causal feature; a classifier $g$ for final classification; a domain-based classifier $g_d$ for invariance goals; and a domain learner $s$ to provide the partition of attack domain. We also provide the algorithm in Appendix~\ref{apd:algo}.

Through theoretical analysis in Proposition~\ref{theo:achiev}, IDEA produces causally invariant defenders under the linear assumption of causal relationship~\cite{arjovsky2019invariant}, enabling graph adversarial robustness.

\begin{proposition}
\label{theo:achiev}
Let $Y=C\gamma+\epsilon$ where $\gamma$ is the causal mechanism, $\epsilon \sim \mathcal{N}(0, \sigma^2)$ is Gaussian noise. 
Let $\rho(\hat{G})=\psi(C, N)$ where $\psi$ is the mapping from causal feature $C$ and non-causal feature $N$ to graph representation $\rho(\hat{G})$, and $\rho$ is a powerful graph representation extractor can extract all information from $\hat{G}$.
Encoder $\Phi$ comprises $\rho$ and a learner $\phi$ with parameter $\Theta_{\phi}$ to learn $C$.
Suppose a function $\tilde{\psi}$ satisfying $\tilde{\psi}(\rho(\hat{G}))=C$, with parameters $\Theta_{\tilde{\psi}}$.
Assume the rank of $\Theta_\phi$ is $r$.
Let $\Theta_{\tilde{\psi}}^\top \Theta_{\gamma}$ and $\Theta_{\phi}^\top \Theta_\omega$ be the parameter of the ground truth defender and learned defender.
If $\Theta_{\phi}^\top \Theta_\omega$ satisfies the following conditions in training attack domain set $\mathcal{D}_{\text{tr}}$:\\
(1) {\small$I\left(\Phi(\hat{G}), Y\right)- \left[I\left(Y, D \mid \Phi(\hat{G})\right)+ I\left(Y, D \mid \Phi(\hat{G})_{\mathcal{N}}\right)\right]$} (Eq.~\ref{eq:goal}) is maximized, \\
%(2) {\small$\left\{\mathbb{E}_{\rho(\hat{G}^{D})}\left[\rho(\hat{G}^{D}) \rho(\hat{G}^D)^{\top}\right]
%\left(\Theta_{\phi}^\top \Theta_\omega - \Theta_{\tilde{\psi}}^\top \Theta_\gamma\right)\right\}_{D\in \mathcal{D}_{\text{tr}}}$} is linearly independent and 
%{\scriptsize $\operatorname{dim}\left(\operatorname{span}\left(\left\{ \mathbb{E}_{\rho(\hat{G})_{i}}\left[\rho(\hat{G})_i \rho(\hat{G})_i^{ \top}\right]\left(\Theta_{\phi}^\top \Theta_\omega - \Theta_{\tilde{\psi}}^\top \Theta_\gamma \right)\right\}_{i\in \mathcal{V}}\right)\right)>\operatorname{dim}(\phi) -r$}, \\
%then $\Theta_{\phi}^\top \Theta_\omega=\Theta_{\tilde{\psi}}^\top \Theta_\gamma$ is causal invariant defender for all attack domain set $\mathcal{D}_{\text{all}}$.
(2) Consider the set $\{\mathbb{E}_{\rho(\hat{G}^{D})}[\rho(\hat{G}^{D}) \rho(\hat{G}^D)^{\top}](\Theta_{\phi}^\top \Theta_\omega - \Theta_{\tilde{\psi}}^\top \Theta_\gamma)\}_{D\in \mathcal{D}_{\text{tr}}}$, which we will refer to as $\mathcal{S}$. If $\mathcal{S}$ is linearly independent and the dimension of the span of $\mathcal{S}$, denoted as $\operatorname{dim}(\operatorname{span}(\mathcal{S}))$, exceeds $\operatorname{dim}(\phi) -r$, then the equality $\Theta_{\phi}^\top \Theta_\omega=\Theta_{\tilde{\psi}}^\top \Theta_\gamma$ defines a causal invariant defender for all attack domain sets $\mathcal{D}_{\text{all}}$.
\end{proposition}
The proof of Proposition~\ref{theo:achiev} is available in Appendix~\ref{proof:achiev}.  
Condition (1) aligns with minimizing the losses $\mathcal{L}_\mathcal{P}$, $\mathcal{L}_\mathcal{I}$, and $\mathcal{L}_\mathcal{E}$ in Eq.\ref{eq:final_loss}. In condition (2), the first term corresponds to minimizing $\mathcal{L}_\mathcal{D}$ in Eq.\ref{eq:final_loss}, while the second term ensures diversity of adversarial examples, common in graph.  
Proposition~\ref{theo:achiev} serves as a theoretical validation for the effectiveness of IDEA.

% Condition (1) can be achieved by minimizing the losses $\mathcal{L}_\mathcal{P}$, $\mathcal{L}_\mathcal{I}$, and $\mathcal{L}_\mathcal{E}$ in Eq.\ref{eq:final_loss}. 
% The first term in condition (2) can be achieved by minimizing $\mathcal{L}_\mathcal{D}$ in Eq.\ref{eq:final_loss}.
% The second term implies the diversity of adversarial examples. 
% % 
% Therefore, IDEA seeks an invariant defender, with proof in Appendix A.2.% ~\ref{proof:achiev}.
% % The proof of Proposition~\ref{theo:achiev} is in Appendix~\ref{proof:achiev}.

\section{Experiments}

%In this section, we conduct extensive experiments under both evasion and poisoning attacks on five public datasets to evaluate the effectiveness of our IDEA method. 
%\subsection{Experimental Settings}

\subsection{Experimental Settings}
%\textbf{Datasets.}
%To evaluate the adaptability of IDEA across various datasets, we conduct node classification experiments on 5 diverse network benchmarks. These include three citation networks: Cora~\cite{bojchevski2019certifiable}, Citeseer~\cite{bojchevski2019certifiable}, and obgn-arxiv~\cite{hu2020open}, a social network Reddit~\cite{hamilton2017inductive,zeng2019graphsaint}, as well as a product co-purchasing network ogbn-products~\cite{hu2020open}. The statistics of datasets are in Appendix D.1.

\subsubsection{Datasets}
\label{apd:dataset}

% \begin{wraptable}{R}{0.49\textwidth}
\begin{table}[t]
\centering
\caption{Statistics of benchmark datasets}
\label{tab:dataset}
\resizebox{0.65\textwidth}{!}{
\begin{tabular}{l|c|rrrrr}
\toprule
Dataset & Type & \#Nodes & \#Edges  & \#Attr. & Classes \\
\midrule
Cora & Citation network & 2,485 & 5,069 & 1,433 & 7 \\
Citeseer & Citation network & 2,110 & 3,668 & 3,703 & 6  \\
Reddit & Social network & 10,004 & 73,512 & 602 & 41\\
ogbn-products & Co-purchasing network & 10,494 & 38,872 & 100 & 35  \\
ogbn-arxiv & Citation network & 169,343 & 2,484,941  & 128 & 39 \\
\bottomrule
\end{tabular}
%\footnotesize{$^*$ $N_{LCC}$ denotes the number of nodes in LCC, $d$ is the dimension of node features, $K$ is the dimension of node label, and $D_{avg}$ denotes the average degree of nodes.}\\
}
\end{table}

We conduct node classification experiments on 5 diverse network benchmarks:  three citation networks (Cora~\cite{Jin2020GraphSL}, Citeseer~\cite{Jin2020GraphSL}, and obgn-arxiv~\cite{hu2020open}), a social network (Reddit~\cite{hamilton2017inductive, zeng2019graphsaint}), and a product co-purchasing network (ogbn-products~\cite{hu2020open}).
Due to the high complexity of some GNN and defense methods, it is difficult to apply them to very large graphs with more than million nodes. Thus, we utilize subgraphs from Reddit and ogbn-products for experiments. Following the settings of most methods~\cite{zugner2018adversarial,zugner_adversarial_2019,TaoGNIA,Jin2020GraphSL,jin2021node,liu2021elastic,li2022reliable}, experiments are conducted on the largest connected component (LCC). All datasets can be assessed at \url{https://anonymous.4open.science/r/IDEA}.
The statistics of datasets are summarized in Table~\ref{tab:dataset}.
\begin{itemize}[leftmargin=10pt]
   \item \emph{Cora}~\cite{Jin2020GraphSL}: A node represents a paper with key words as attributes and paper class as label, and the edge represents the citation relationship.
   \item \emph{Citeseer}~\cite{Jin2020GraphSL}: Same as \emph{Cora}.
   \item \emph{Reddit}~\cite{hamilton2017inductive,zeng2019graphsaint}: Each node represents a post, with word vectors as attributes and community as the label, while each edge represents the post-to-post relationship. 
   \item \emph{ogbn-products}~\cite{hu2020open}: A node represents a product sold in Amazon with the word vectors of product descriptions as attributes and the product category as the label, and edges between two products indicate that the products are purchased together. 
   \item \emph{ogbn-arxiv}~\cite{hu2020open}: Each node denotes a Computer Science (CS) arXiv paper indexed by ~\cite{wang2020microsoft} with attributes obtained by averaging the embeddings of words in paper's title and abstract. Each edge indicates the citation relationship, and the node label is the primary categories of each arXiv paper. 
\end{itemize}
%~\ref{apd:dataset}.
%Experiments are performed on each dataset's largest connected component (LCC) following existing works~\cite{zugner2018adversarial,zugner_adversarial_2019,liu2021elastic,Jin2020GraphSL}, with statistics summarized in Table~\ref{tab:dataset}. 

\subsubsection{Defense baselines}
\label{apd:baseline}
We evaluate the performance of our proposed method, IDEA, by comparing it against ten baseline approaches. These baselines include traditional Graph Neural Networks (GNNs) and defense techniques from three main categories: graph purification, robust aggregation, and adversarial training. For each category, we select the most representative and state-of-the-art methods for comparison. In summary, our comparison includes the following ten baselines:
\begin{itemize}[leftmargin=10pt]
   \item Traditional GNNs 
   \begin{enumerate}
   \item \emph{GCN}~\cite{kipf2017semi}: GCN is a popular graph convolutional network based on spectral theory. 
   \item \emph{GAT}~\cite{velickovic2018graph}: GAT computes the hidden representations of each node by attending over its neighbors via graph attentional layers.
   \end{enumerate}
   \item Graph purification
   \begin{enumerate}[resume*]
       \item \emph{ProGNN}~\cite{Jin2020GraphSL}: ProGNN simultaneously learns the graph structure and GNN parameters by optimizing three regularizations, i.e., feature smoothness, low-rank and sparsity. 
       \item \emph{STABLE}~\cite{li2022reliable}: STABLE first learns reliable representations of graph structure via unsupervised learning, and then designs an advanced GCN as a downstream classifier to enhance the robustness of GCN.
       \item \emph{GARNET}~\cite{GARNET22DengLF}: GARNET uses weighted spectral embedding to create a base graph, then refines this graph through the pruning of non-essential edges to enhance adversarial robustness.
   \end{enumerate}
   \item Robust aggregation
   \begin{enumerate}[resume*]
   \item \emph{RGCN}~\cite{Zhu2019RobustGC}: RGCN uses gaussian distributions in graph convolutional layers to absorb the effects of adversarial attacks.
   \item \emph{SimPGCN}~\cite{jin2021node}: SimPGCN presents a feature similarity preserving aggregation which balances the structure and feature information, and self-learning regularization to capture the feature similarity and dissimilarity between nodes.
   \item \emph{Elastic}~\cite{liu2021elastic}: Elastic enhances the local smoothness adaptivity of GNNs via $\ell_1$-based graph smoothing and derives the elastic message passing (EMP).
   \item \emph{Soft-Median}~\cite{GeislerSSZBG21}: Soft-Median is robust aggregation function where the weight for each instance is determined based on the distance to the dimension-wise median. 
   \end{enumerate}
   \item Adversarial training
   \begin{enumerate}[resume*]
   \item \emph{FLAG}~\cite{kong2020flag}: FLAG, a state-of-the-art adversarial training method, defends against attacks by incorporating adversarial examples into the training set, enabling the model to correctly classify them.
   \end{enumerate}
\end{itemize}

\subsubsection{Attack Methods}
\label{apd:atk}
We assess the robustness of IDEA by examining its performance against five adversarial attacks, including one representative poisoning attack (MetaAttack~\cite{zugner_adversarial_2019}) and four evasion attacks (nettack~\cite{zugner2018adversarial}, PGD~\cite{Madry2017TowardsDL}, TDGIA~\cite{ZouTDGIA}, G-NIA~\cite{TaoGNIA}). Among these attacks, nettack and MetaAttack modify the original graph structure, while PGD, TDGIA, and G-NIA are node injection attacks. The following is a brief description of each attack:
\begin{itemize}[leftmargin=10pt]
   \item \emph{nettack}~\cite{zugner2018adversarial}: Nettack is the first adversarial attack on graph data,  which can attack node attributes and graph structure with gradient. In this paper, we adopt nettack to attack graph structure, i.e., adding and removing edges.
   \item \emph{PGD}~\cite{Madry2017TowardsDL}: PGD, a popular adversarial attack, is used as node injection attack. We employi projected gradient descent (PGD) to inject malicious nodes on graphs.
   \item \emph{TDGIA}~\cite{ZouTDGIA}: TDGIA consists of two modules: the heuristic topological defective edge selection for injecting nodes and smooth adversarial optimization for generating features of injected nodes.
   \item \emph{G-NIA}~\cite{TaoGNIA}: G-NIA is  one of the state-of-the-art node injection attack methods, showing excellent attack performance. G-NIA models the optimization process via a parametric model to preserve the learned attack strategy and reuse it when inferring.
   \item \emph{MetaAttack}~\cite{zugner_adversarial_2019}: MetaAttack is the most representative poisoning attack method, which has been widely-used to evaluate the robustness of GNN models. 
\end{itemize}

\subsubsection{Implementation Details}
\label{apd:imp}
For each dataset, we randomly split nodes as 1:1:8 for training, validation and test, following~\cite{Jin2020GraphSL,jin2021node,liu2021elastic,li2022reliable}.
For each experiment, we report the average performance and the standard deviation of 10 runs.
For attack and defense methods, we employ the most widely recognized DeepRobust~\cite{deeprobust} benchmark in the field of graph adversarial and defense,  to ensure that the experimental results can be compared directly to other papers that use DeepRobust (such as Elastic~\cite{liu2021elastic}, ProGNN~\cite{Jin2020GraphSL}, STABLE~\cite{li2022reliable}, and SimPGCN~\cite{jin2021node}). 
%We employ the widely-used DeepRobust~\cite{deeprobust} library for the attack and defense methods. We tune their hyper-parameters according validation set. 
Note that MetaAttack is untargeted attack, performance is reported on the test set with perturbation rates from 0\% to 20\%, following~\cite{liu2021elastic,li2022reliable}. The evasion attacks are targeted attacks, and we randomly sample $20\%$ of all nodes from the test set as targets. Nettack perturbs 20\% edges, while node injection attacks (PGD, TDGIA, and G-NIA) inject 20\% nodes and edges. 
We focused on gray-box attack scenarios following ~\cite{zugner2018adversarial,zugner_adversarial_2019,li2022reliable,jin2021node,Jin2020GraphSL} and the attacker is only aware of the input and output, which is practical.

% Especially, for the graph purification methods under evasion attacks, we first purify the perturbed graph and learn the high-quality representation to take advantage of their performance. For example, STABLE first purifies the graph structure and learns the high-quality representation for each node, then learn a GCN to predict labels. While ProGNN  
For our IDEA, we tune the hyper-parameters from the following range: the coefficient $\alpha$ over $\{10, 25, 100, 150\}$, the number of domains over $\{2, 5, 10, 20\}$. 
The hyper-parameters of all datasets can be assessed at \url{https://anonymous.4open.science/r/IDEA}. Note that we implement both attribute and structural attacks to generate adversarial examples that minimize the predictive loss $\mathcal{L}_\mathcal{P}$. 
Specifically, attribute attack generation is the same as that in FLAG~\cite{kong2020flag}, while structural attack generation is the same as that in EERM~\cite{wu2022handling}.
For all methods that require a backbone model (e.g., FLAG and our IDEA), we use GCN as the backbone model.
All experiments are conducted on a single NVIDIA V100 32 GB GPU.

\begin{table*}[t]
% %%%%%%% \vspace{-10pt}
\caption{Accuracy(\%) of targets under evasion attacks. The \textbf{best} and {\ul second-best} are highlighted. Parentheses denote IDEA's relative increase compared to second-best. ``-'' for out-of-memory (OOM). }
\resizebox{1.02\textwidth}{!}{
\begin{tabular}{ccccccccccccc}
\toprule   
\textbf{Dataset}                        & \textbf{Attack} & \textbf{GCN} & \textbf{GAT} & \textbf{ProGNN} & \textbf{STABLE} & \textbf{GARNET} & \textbf{RGCN} & \textbf{SimpGCN} & \textbf{Elastic}& \textbf{Soft-Median} & \textbf{FLAG}  &\textbf{IDEA} \\
\midrule
\multirow{6}{*}{\textbf{Cora}}          & Clean      & 85.0               $\pm$ 0.5 & 84.6               $\pm$ 0.8 & 81.9              $\pm$ 1.2 & 83.9                $\pm$ 0.6 & 84.2$\pm$0.8 & 83.6                $\pm$ 0.7 & 83.0                 $\pm$ 1.2 & 83.4                $\pm$ 1.9 & 84.3$\pm$0.9 & {\ul 85.8}          $\pm$ 0.6 & \textbf{88.4}        $\pm$ 0.6        \ (\ $\uparrow$ 3.0\%) \\
& nettack         & 83.0               $\pm$ 0.5         & 81.7               $\pm$ 0.7         & 79.9              $\pm$ 1.1             & 21.5                $\pm$ 4.8     &83.0$\pm$1.0       & 81.7                $\pm$ 0.6         & 82.0                 $\pm$ 1.2           & 79.4                $\pm$ 1.7           &83.5$\pm$0.8      & {\ul 84.8}          $\pm$ 0.6       & \textbf{85.4}        $\pm$ 0.7      \ (\ $\uparrow$ 0.8\%) \\
& PGD             & 44.2               $\pm$ 3.4         & 26.7               $\pm$ 7.6         & 19.6              $\pm$ 2.2             & 32.2                $\pm$ 0.2       &{\ul 81.4}$\pm$1.6    & 80.5          $\pm$ 0.4         & 9.0                  $\pm$ 2.2           & 29.0                $\pm$ 5.5           & 19.5$\pm$2.3         & 60.2                $\pm$ 2.4    & \textbf{83.6}        $\pm$ 2.1    \ (\ $\uparrow$ 2.6\%) \\
& TDGIA           & 20.2               $\pm$ 2.3         & 33.7               $\pm$ 14.9        & 15.4              $\pm$ 1.7             & 30.9                $\pm$ 2.4        &76.1$\pm$2.9     & {\ul 79.9}          $\pm$ 0.9         & 13.5                 $\pm$ 1.2           & 18.0                $\pm$ 1.2           &  16.8$\pm$0.4      & 57.2                $\pm$ 3.0    & \textbf{81.2}        $\pm$ 2.5    \ (\ $\uparrow$ 1.6\%) \\
& G-NIA           & 2.3                $\pm$ 0.5         & 5.2                $\pm$ 3.1         & 4.2               $\pm$ 0.8             & 25.8                $\pm$ 10.1     & 6.1$\pm$0.9      & {\ul 81.3}          $\pm$ 0.9         & 11.5                 $\pm$ 8.1           & 11.2                $\pm$ 3.7          & 4.8$\pm$0.8         & 64.8                $\pm$ 2.0     & \textbf{85.3}        $\pm$ 1.2        \ (\ $\uparrow$ 4.9\%) \\
\cmidrule(l){2-13}
& AVG             & 47.0               $\pm$ 37.0          & 46.4               $\pm$ 35.2          & 40.2              $\pm$ 37.6              & 38.9                $\pm$ 25.5      & 66.2$\pm$33.7      & {\ul 81.4}          $\pm$ 1.4          & 39.8                 $\pm$ 39.0            & 44.2                $\pm$ 34.6         & 41.8$\pm$38.8       & 70.6                $\pm$ 13.7          & \textbf{84.8}        $\pm$ 2.7         \ (\ $\uparrow$ 4.1\%) \\
\midrule
\multirow{6}{*}{\textbf{Citeseer}}      & Clean           & 73.6               $\pm$ 0.6         & 74.7               $\pm$ 1.0         & 74.1              $\pm$ 0.9             & {\ul 75.2}          $\pm$ 0.5     & 71.3$\pm$1.0       & 74.6                $\pm$ 0.5         & 74.9                 $\pm$ 1.3           & 74.0                $\pm$ 1.3         & 73.6$\pm$0.9       & 74.7                $\pm$ 0.9        & \textbf{82.0}        $\pm$ 1.9        \ (\ $\uparrow$ 9.1\%) \\
& nettack         & 72.6               $\pm$ 0.7         & 72.6               $\pm$ 1.8         & 71.5              $\pm$ 0.9             & 20.8                $\pm$ 8.5      &70.3$\pm$1.1      & 73.2                $\pm$ 0.6         & {\ul 74.5}           $\pm$ 1.1           & 71.8                $\pm$ 1.5          & 72.8$\pm$0.9     & 73.6                $\pm$ 1.2         & \textbf{78.8}        $\pm$ 1.6        \ (\ $\uparrow$ 5.7\%) \\
& PGD             & 52.7               $\pm$ 4.5         & 54.5               $\pm$ 5.3         & 41.4              $\pm$ 4.1             & 17.7                $\pm$ 6.2      &65.4$\pm$1.1      & {\ul 70.1}          $\pm$ 1.1         & 48.2                 $\pm$ 13.9          & 39.1                $\pm$ 6.0            & 36.2$\pm$1.9   & 60.1                $\pm$ 2.5         & \textbf{76.9}        $\pm$ 3.4        \ (\ $\uparrow$ 9.8\%) \\
& TDGIA           & 23.0               $\pm$ 3.8         & 44.7               $\pm$ 11.2        & 16.9              $\pm$ 2.1             & 15.5                $\pm$ 5.3       & 57.1$\pm$2.4    & {\ul 63.8}          $\pm$ 7.4         & 28.1                 $\pm$ 11.1          & 18.2                $\pm$ 3.6         & 21.6$\pm$1.1       & 57.5                $\pm$ 1.7        & \textbf{75.9}        $\pm$ 3.9        \ (\ $\uparrow$ 19.0\%) \\
& G-NIA           & 15.0               $\pm$ 3.6         & 13.6               $\pm$ 3.6         & 22.5              $\pm$ 4.8             & 18.5                $\pm$ 6.6     &18.2$\pm$0.8       & 32.1                $\pm$ 6.4         & 54.4                 $\pm$ 16.8          & 30.2                $\pm$ 4.2        & 14.1$\pm$1.0        & {\ul 68.0}          $\pm$ 0.9        & \textbf{79.4}        $\pm$ 3.0        \ (\ $\uparrow$ 16.8\%) \\
\cmidrule(l){2-13}
& AVG             & 47.4               $\pm$ 27.3          & 52.0               $\pm$ 24.9          & 45.3              $\pm$ 26.7              & 29.5                $\pm$ 25.6    &56.4$\pm$22.1        & 62.8                $\pm$ 17.7          & 56.0                 $\pm$ 19.6            & 46.6                $\pm$ 25.1            & 43.6$\pm$28.1    & {\ul 66.8}          $\pm$ 7.8           & \textbf{78.6}        $\pm$ 2.4         \ (\ $\uparrow$ 17.7\%) \\
\midrule
\multirow{6}{*}{\textbf{Reddit}}        & Clean           & 84.9               $\pm$ 0.6         & {\ul 88.5}         $\pm$ 0.3         & 66.2              $\pm$ 3.1             & 83.6                $\pm$ 0.4       & 85.7$\pm$0.3     & 68.0                $\pm$ 1.7         & 50.2                 $\pm$ 8.3           & 72.7                $\pm$ 0.6         & 85.6$\pm$0.8       & 86.9                $\pm$ 0.4        & \textbf{90.8}        $\pm$ 0.3        \ (\ $\uparrow$ 2.7\%) \\
& nettack         & 84.8               $\pm$ 0.5         & {\ul 87.9}         $\pm$ 0.4         & 68.8              $\pm$ 3.1             & 3.6                 $\pm$ 3.2       & 86.5$\pm$0.4    & 67.0                $\pm$ 1.7         & 49.5                 $\pm$ 8.4           & 71.4                $\pm$ 0.7           &  84.5$\pm$0.8     & 85.5                $\pm$ 0.4       & \textbf{89.1}        $\pm$ 0.5        \ (\ $\uparrow$ 1.4\%) \\
& PGD             & 46.0               $\pm$ 1.6         & 30.8               $\pm$ 2.5         & 20.0              $\pm$ 5.4             & 3.6                 $\pm$ 1.5     & {\ul 81.2}$\pm$0.8      & 53.1                $\pm$ 2.1         & 9.8                  $\pm$ 3.3           & 19.0                $\pm$ 1.0           & 16.5$\pm$0.9     &  72.1          $\pm$ 0.9        & \textbf{81.6}        $\pm$ 0.9        \ (\ $\uparrow$ 0.4\%) \\
& TDGIA           & 24.1               $\pm$ 1.6         & 32.8               $\pm$ 3.8         & 9.0               $\pm$ 3.0             & 3.6                 $\pm$ 1.4       & 48.5$\pm$1.5     & 44.3                $\pm$ 1.9         & 5.5                  $\pm$ 1.6           & 8.3                 $\pm$ 0.6            & 6.3$\pm$0.7    & {\ul 73.1}          $\pm$ 0.7       & \textbf{81.3}        $\pm$ 0.6        \ (\ $\uparrow$ 11.1\%) \\
& G-NIA           & 1.0                $\pm$ 0.8         & 2.5                $\pm$ 1.1         & 4.0               $\pm$ 3.6             & 4.7                 $\pm$ 2.2        &8.8$\pm$2.6     & 5.0                 $\pm$ 2.0         & 5.3                  $\pm$ 3.7           & 3.3                 $\pm$ 0.7            & 1.9$\pm$0.9   & {\ul 76.9}          $\pm$ 1.2        & \textbf{84.2}        $\pm$ 1.1        \ (\ $\uparrow$ 9.5\%) \\
\cmidrule(l){2-13}
& AVG             & 48.2               $\pm$ 37.1          & 48.5               $\pm$ 38.2          & 33.6              $\pm$ 31.5              & 19.8                $\pm$ 35.6     & 62.1$\pm$33.7       & 47.5                $\pm$ 25.7          & 24.1                 $\pm$ 23.6            & 34.9                $\pm$ 34.4           &  39.0$\pm$42.4        & {\ul 78.9}          $\pm$ 6.9      & \textbf{85.4}        $\pm$ 4.4         \ (\ $\uparrow$ 8.2\%) \\
\midrule
\multirow{6}{*}{\tabincell{c}{\textbf{ogbn-}\\ \textbf{products}}} & Clean           & 63.9               $\pm$ 0.7         &  69.6         $\pm$ 0.4         & 49.7              $\pm$ 2.4             & 67.6                $\pm$ 0.8  &71.9$\pm$0.6          & 64.3                $\pm$ 0.4         & 57.7                 $\pm$ 2.2           & 57.9                $\pm$ 0.9            & {\ul 72.8}$\pm$0.4      & 67.6                $\pm$ 0.7     & \textbf{76.1}        $\pm$ 0.4        \ (\ $\uparrow$ 4.5\%) \\
& nettack         & 63.3               $\pm$ 0.6         & 62.1               $\pm$ 2.1         & 50.1              $\pm$ 3.1             & 14.9                $\pm$ 1.4      & {\ul 72.8}$\pm$0.7      & 62.1                $\pm$ 0.8         & 56.1                 $\pm$ 2.4           & 52.6                $\pm$ 0.9            & 71.3$\pm$0.5    & 65.8          $\pm$ 0.5        & \textbf{74.4}        $\pm$ 0.6        \ (\ $\uparrow$ 2.3\%) \\
& PGD             & 32.2               $\pm$ 0.9         & 25.0               $\pm$ 0.8         & 17.5              $\pm$ 1.0             & 14.3                $\pm$ 1.8       & {\ul 57.7}$\pm$2.7     & 34.8                $\pm$ 0.9         & 17.8                 $\pm$ 1.5           & 21.2                $\pm$ 0.5         &   19.8$\pm$0.7      & 54.0          $\pm$ 0.6       & \textbf{67.9}        $\pm$ 0.6        \ (\ $\uparrow$ 17.5\%) \\
& TDGIA           & 23.1               $\pm$ 1.0         & 16.9               $\pm$ 1.1         & 11.0              $\pm$ 0.5             & 16.5                $\pm$ 1.9       & {\ul 54.6}$\pm$2.4     & 28.0                $\pm$ 0.9         & 16.5                 $\pm$ 2.2           & 16.9                $\pm$ 0.8              & 11.7$\pm$0.5  & 49.5          $\pm$ 1.0       & \textbf{64.9}        $\pm$ 0.9        \ (\ $\uparrow$ 18.8\%) \\
& G-NIA           & 2.7                $\pm$ 1.0         & 3.6                $\pm$ 2.0         & 2.2               $\pm$ 0.8             & 9.9                 $\pm$ 5.2       &4.4$\pm$1.0    & 7.1                 $\pm$ 2.6         & 8.6                  $\pm$ 4.5           & 3.3                 $\pm$ 0.6            &  0.8$\pm$0.3   & {\ul 54.2}          $\pm$ 0.8        & \textbf{65.6}        $\pm$ 1.1        \ (\ $\uparrow$ 21.0\%) \\
\cmidrule(l){2-13}
& AVG             & 37.1               $\pm$ 26.5          & 35.5               $\pm$ 28.9          & 26.1              $\pm$ 22.4              & 24.7                $\pm$ 24.1         & 52.3$\pm$28.0   & 39.3                $\pm$ 24.1          & 31.3                 $\pm$ 23.6            & 30.4                $\pm$ 23.7          & 35.3$\pm$34.2       & {\ul 58.2}          $\pm$ 8.0         & \textbf{69.8}        $\pm$ 5.2         \ (\ $\uparrow$ 19.8\%) \\
\midrule
\multirow{5}{*}{\tabincell{c}{\textbf{ogbn-}\\ \textbf{arxiv}}}    & Clean           & {\ul 65.3}         $\pm$ 0.3         & 65.2               $\pm$ 0.1         & -                                & -                           &53.0$\pm$0.1     & 60.2                $\pm$ 1.0         & -                                 & 58.0                $\pm$ 0.1      & 61.1$\pm$0.2           & 61.0                $\pm$ 0.7      & \textbf{66.7}        $\pm$ 0.4        \ (\ $\uparrow$ 2.1\%) \\
& PGD             & 41.1         $\pm$ 1.0         & 22.6               $\pm$ 1.9         & -                                & -                          & {\ul 52.3}$\pm$0.1       & 37.8                $\pm$ 2.0         & -                                 & 29.9                $\pm$ 0.6          &  19.1$\pm$0.4     & 24.2                $\pm$ 2.8       & \textbf{52.9}        $\pm$ 1.0        \ (\ $\uparrow$ 1.2\%) \\
& TDGIA           &  33.1         $\pm$ 1.6         & 9.7                $\pm$ 1.8         & -                                & -                            & {\ul 51.6}$\pm$0.1     & 27.5                $\pm$ 2.1         & -                                 & 20.5                $\pm$ 0.9        &  18.6$\pm$0.8        & 29.3                $\pm$ 2.3      & \textbf{53.2}        $\pm$ 0.8        \ (\ $\uparrow$ 3.0\%) \\
& G-NIA           & 4.6                $\pm$ 0.4         & 2.5                $\pm$ 0.3         & -                                & -                            & {\ul 35.3}$\pm$0.1    & 5.6                 $\pm$ 0.8         & -                                 & 14.6                $\pm$ 0.1           &  2.5$\pm$0.1    & 11.5          $\pm$ 1.1        & \textbf{40.5}        $\pm$ 1.6        \ (\ $\uparrow$ 14.5\%) \\
\cmidrule(l){2-13}
& AVG             & 36.0         $\pm$ 25.0          & 25.0               $\pm$ 28.1          & -                                & -                         & {\ul 48.1}$\pm$8.5       & 32.8                $\pm$ 22.7          & -                                 & 30.7                $\pm$ 19.2              &  27.3$\pm$22.2    & 31.5                $\pm$ 21.0       & \textbf{53.3}        $\pm$ 10.7     
    \ (\ $\uparrow$ 10.9\%) 
    \\
\bottomrule
\end{tabular}
}
\label{tab:evasion}
% \vspace{-5pt}
\end{table*}

\subsection{Robustness against Evasion Attacks}
We conduct experiments under four evasion attacks (nettack, PGD, TDGIA, and G-NIA), and show the accuracy of target nodes in Table~\ref{tab:evasion}.  
We also report the average accuracy of clean and attacked graphs, along with standard deviation of accuracy across these graphs,  denoted as AVG. Note that we exclude nettack from ogbn-arxiv evaluation due to its lack of scalability.
GCN and GAT exhibit high accuracy on clean graphs, however, their accuracy significantly declines under PGD, TDGIA, and G-NIA. 
% These results highlight the vulnerability of GNNs to adversarial attacks.
Defense methods suffer from severe performance degradation under various attacks, and some (such as ProGNN (81.9\%) and RGCN (83.6\%)) even experience a decline on Clean. For graph purification methods,  ProGNN and STABLE perform poorly under most attacks, maybe because they require retraining to achieve defensive effects, rendering them unsuitable for evasion attacks. GARNET shows effectiveness against PGD and TDGIA, but still struggles to defend against G-NIA.
RGCN, SimPGCN, Elastic, and Soft-Median perform well against nettack; however, they suffer from performance degradation on clean graphs, which is undesirable.
%Regarding adversarial training, the state-of-the-art method FLAG outperforms other baselines, nonetheless, FLAG does not excel on all datasets, for instance, its defense performances on Cora and Citeseer are unsatisfactory.
Adversarial training FLAG outperforms other baselines but exhibits unsatisfactory defense on Cora,  Citeseer, and ogbn-arxiv.

Our proposed method, IDEA, achieves the best performance on Clean and across all attacks, significantly outperforming all baselines on all datasets. 
% On Clean, IDEA shows the best performance, which mainly comes from learning causal features that have strong predictability for labels. Moreover, the performance of IDEA remains consistent under both clean graphs and various attacks, as evidenced by the low standard deviation in AVG. This highlights its invariant predictability across all attacks. For example, on Citeseer, IDEA's standard deviation across graphs is only 2.4, while the runner-up, FLAG, reaches 7.8. These results demonstrate IDEA not only has strong predictability (high accuracy on Clean) but also invariant predictability (sustained accuracy across all attacks).
% emphasizing its robustness and effectiveness.
On Clean, IDEA exhibits the best performance primarily due to its ability to learn causal features that have strong label predictability.  Furthermore, IDEA's performance remains good consistency under both clean graphs and various attacks, evidenced by the low standard deviation in AVG. This emphasizes its invariant predictability across all attacks. For instance, on Citeseer, IDEA's standard deviation across graphs is a only 2.4, while the runner-up, FLAG, reaches 7.8. These results demonstrate that IDEA possesses both strong predictability (high accuracy on Clean) and invariant predictability (sustained accuracy across attacks).

% \begin{table}[]
% %%%%%%% \vspace{-3.5mm}
% \caption{Accuracy(\%) of test set under poisoning attack (MetaAttack). 
% % The top two performance is highlighted in bold and underlined. 
% }
% \resizebox{1.01\textwidth}{!}{
% \begin{tabular}{ccccccccccccc}
%\textbf{Dataset}                        & \textbf{Pb. rate} & \textbf{GCN} & \textbf{GAT} & \textbf{ProGNN} & \textbf{STABLE} & \textbf{GARNET} & \textbf{RGCN} & \textbf{SimpGCN} & \textbf{Elastic} & \textbf{Soft-Median} & \textbf{FLAG} & \textbf{IDEA}      
%\end{tabular}
% \label{tab:poison}
% \end{table}

\begin{table*}[ht]
\caption{Accuracy(\%) of test set under poisoning attack (MetaAttack).}
\label{tab:poison}
\resizebox{\textwidth}{!}{
\begin{tabular}{crccccccccccc}
\toprule
\textbf{Dataset}                        & \textbf{Pb. rate} & \textbf{GCN} & \textbf{GAT} & \textbf{ProGNN} & \textbf{STABLE} & \textbf{GARNET} & \textbf{RGCN} & \textbf{SimpGCN} & \textbf{Elastic} & \textbf{Soft-Median} & \textbf{FLAG} & \textbf{IDEA} \\
\midrule
\multirow{5}{*}{\textbf{Cora}}          & \textbf{0\%}      & 83.6$\pm$0.5     & 83.5$\pm$0.5     & 83.0$\pm$0.2          & 85.6$\pm$0.6        & 80.1$\pm$0.5        & 82.6$\pm$0.3      & 81.9$\pm$1.0          & {\ul85.8}$\pm$0.4            & 84.0$\pm$0.5               & 83.4$\pm$0.3      & \textbf{87.1}$\pm$0.7    \ (\ $\uparrow$ 1.5\%)  \\
                                       & \textbf{5\%}      & 77.8$\pm$0.6     & 80.3$\pm$0.5     & {\ul82.3}$\pm$0.5        & 81.4$\pm$0.5        & 77.1$\pm$0.8        & 77.5$\pm$0.5      & 77.6$\pm$0.7         & 82.2$\pm$0.9            & 79.9$\pm$0.8             & 80.9$\pm$0.3      & \textbf{85.5}$\pm$0.6    \ (\ $\uparrow$ 3.9\%)    \\
                                       & \textbf{10\%}     & 74.9$\pm$0.7     & 78.5$\pm$0.6     & 79.0$\pm$0.6          & {\ul80.5}$\pm$0.6        & 75.3$\pm$0.8        & 73.7$\pm$1.2      & 75.7$\pm$1.1         & 78.8$\pm$1.7            & 73.4$\pm$2.3             & 78.8$\pm$0.9      & \textbf{84.8}$\pm$0.6    \ (\ $\uparrow$ 5.3\%)  \\
                                       & \textbf{15\%}     & 67.8$\pm$1.2     & 73.6$\pm$0.8     & 76.4$\pm$1.3        & {\ul78.6}$\pm$0.4        & 72.3$\pm$0.7        & 70.2$\pm$0.6      & 72.7$\pm$2.8         & 77.2$\pm$1.6            & 70.5$\pm$1.1             & 75.0$\pm$0.7        & \textbf{84.2}$\pm$0.6   \ (\ $\uparrow$ 7.1\%)   \\
                                       & \textbf{20\%}     & 61.6$\pm$1.1     & 66.6$\pm$0.8     & 73.3$\pm$1.6        & {\ul77.8}$\pm$1.1        & 69.8$\pm$0.7        & 62.7$\pm$0.7      & 70.3$\pm$4.6         & 70.5$\pm$1.3            & 60.5$\pm$0.4             & 70.2$\pm$1.1      & \textbf{83.2}$\pm$0.6     \ (\ $\uparrow$ 6.9\%) \\
\midrule
\multirow{5}{*}{\textbf{Citeseer}}      & \textbf{0\%}      & 73.3$\pm$0.3     & 74.4$\pm$0.8     & 73.3$\pm$0.7        & {\ul75.8}$\pm$0.4        & 70.4$\pm$0.7        & 74.4$\pm$0.3      & 74.4$\pm$0.7         & 73.8$\pm$0.6            & 71.3$\pm$0.8             & 72.8$\pm$0.8      & \textbf{80.3}$\pm$1.0   \ (\ $\uparrow$ 5.9\%)    \\
                                       & \textbf{5\%}      & 70.2$\pm$0.8     & 72.3$\pm$0.5     & 72.9$\pm$0.6        & {\ul74.1}$\pm$0.6        & 69.2$\pm$0.9        & 71.7$\pm$0.3      & 73.3$\pm$1.0          & 72.9$\pm$0.5            & 69.6$\pm$2.2             & 71.1$\pm$0.6      & \textbf{79.1}$\pm$0.9    \ (\ $\uparrow$ 6.7\%)  \\
                                       & \textbf{10\%}     & 68.0$\pm$1.4       & 70.3$\pm$0.7     & 72.5$\pm$0.8        & {\ul73.5}$\pm$0.4        & 68.5$\pm$1.0          & 69.3$\pm$0.4      & 72.0$\pm$1.0            & 72.6$\pm$0.4            & 67.9$\pm$1.9             & 69.2$\pm$0.6      & \textbf{78.6}$\pm$1.1    \ (\ $\uparrow$ 6.9\%)  \\
                                       & \textbf{15\%}     & 65.2$\pm$0.9     & 67.7$\pm$1.0       & 72.0$\pm$1.1          & {\ul73.2}$\pm$0.5        & 65.0$\pm$1.2          & 66.0$\pm$0.2        & 70.8$\pm$1.3         & 71.9$\pm$0.7            & 66.0$\pm$2.9               & 66.5$\pm$0.8      & \textbf{77.6}$\pm$0.6    \ (\ $\uparrow$ 6.0\%)  \\
                                       & \textbf{20\%}     & 60.1$\pm$1.4     & 64.3$\pm$1.0       & 70.0$\pm$2.3          & {\ul72.8}$\pm$0.5        & 62.9$\pm$1.9        & 61.2$\pm$0.5      & 70.0$\pm$1.7           & 64.7$\pm$0.8            & 56.1$\pm$1.3             & 64.1$\pm$0.8      & \textbf{77.8}$\pm$0.9    \ (\ $\uparrow$ 6.9\%)  \\
\midrule
\multirow{5}{*}{\textbf{Reddit}}        & \textbf{0\%}      & 84.5$\pm$0.5     & 88.0$\pm$0.3       & 73.4$\pm$2.8        & 86.6$\pm$0.2        & 85.2$\pm$0.2        & 78.2$\pm$0.6      & 51.4$\pm$7.6         & 83.8$\pm$0.3            & {\ul88.8}$\pm$0.5             & 84.6$\pm$0.2      & \textbf{91.2}$\pm$0.3   \ (\ $\uparrow$ 2.7\%)   \\
                                       & \textbf{5\%}      & 81.0$\pm$0.8       & {\ul86.1}$\pm$0.6     & 73.9$\pm$1.2        & 81.5$\pm$0.4        & 78.3$\pm$0.3        & 73.9$\pm$1.7      & 34.8$\pm$9.9         & 80.6$\pm$0.5            & 82.6$\pm$0.9             & 83.6$\pm$0.4      & \textbf{90.0}$\pm$0.4    \ (\ $\uparrow$ 4.5\%)  \\
                                       & \textbf{10\%}     & 72.1$\pm$0.6     & {\ul78.8}$\pm$0.8     & 63.2$\pm$1.2        & 75.9$\pm$0.5        & 63.4$\pm$1.3        & 57.8$\pm$1.3      & 27.3$\pm$7.8         & 70.4$\pm$1.0             & 66.4$\pm$1.4             & 72.2$\pm$0.8      & \textbf{89.0}$\pm$0.4    \ (\ $\uparrow$ 12.9\%)  \\
                                       & \textbf{15\%}     & 70.1$\pm$1.7     & {\ul76.0}$\pm$1.6       & 59.9$\pm$1.4        & 73.8$\pm$0.4        & 62.2$\pm$0.6        & 53.3$\pm$1.4      & 25.0$\pm$6.4           & 68.7$\pm$0.6            & 60.5$\pm$2.4             & 66.3$\pm$1.2      & \textbf{88.4}$\pm$0.4    \ (\ $\uparrow$ 16.3\%)  \\
                                       & \textbf{20\%}     & 67.9$\pm$1.4     & {\ul72.7}$\pm$1.9     & 56.7$\pm$0.9        & 71.7$\pm$0.5        & 60.7$\pm$0.8        & 51.5$\pm$2.8      & 19.0$\pm$4.5           & 67.4$\pm$0.6            & 56.6$\pm$1.5             & 63.7$\pm$0.5      & \textbf{88.1}$\pm$0.3    \ (\ $\uparrow$ 21.2\%)  \\
\midrule
\multirow{5}{*}{\tabincell{c}{\textbf{ogbn-}\\ \textbf{products}}}& \textbf{0\%}      & 63.0$\pm$0.7       & 68.6$\pm$0.4     & 64.3$\pm$2          & 70.5$\pm$0.5        & 71.3$\pm$0.5        & 63.0$\pm$0.5        & 57.1$\pm$2.1         & 72.7$\pm$0.2            & {\ul74.3}$\pm$0.3             & 66.3$\pm$0.4      & \textbf{75.2}$\pm$0.3    \ (\ $\uparrow$ 1.2\%)  \\
                                       & \textbf{5\%}      & 49.6$\pm$0.9     & 64.4$\pm$0.3     & 50.0$\pm$2.4          & 58.7$\pm$0.7        & 61.5$\pm$0.6        & 40.0$\pm$1.1        & 31.9$\pm$10.3        & 61.0$\pm$0.7              & {\ul66.5}$\pm$0.4             & 57.3$\pm$0.7      & \textbf{73.7}$\pm$0.5    \ (\ $\uparrow$ 10.8\%)  \\
                                       & \textbf{10\%}     & 39.4$\pm$1.1     & 54.4$\pm$0.7     & 43.4$\pm$1.9        & 50.5$\pm$0.5        & 53.6$\pm$0.9        & 33.4$\pm$0.9      & 26.7$\pm$8.1         & 52.3$\pm$0.7            & {\ul56.3}$\pm$0.7             & 46.7$\pm$0.7      & \textbf{72.9}$\pm$0.3    \ (\ $\uparrow$ 29.5\%)  \\
                                       & \textbf{15\%}     & 34.4$\pm$1.1     & 46.7$\pm$0.7     & 38.4$\pm$1.8        & 44.6$\pm$0.8        & 47.7$\pm$0.4        & 29.9$\pm$1.0       & 20.3$\pm$6.6         & {\ul48.6}$\pm$0.4            & 47.7$\pm$1.0               & 42.7$\pm$0.5      & \textbf{71.9}$\pm$0.7     \ (\ $\uparrow$ 47.9\%)   \\
                                       & \textbf{20\%}     & 31.2$\pm$1.0       & 41.9$\pm$1.2     & 34.0$\pm$2.5          & 40.3$\pm$0.7        & 42.3$\pm$0.4        & 27.8$\pm$0.8      & 16.0$\pm$2.1           & {\ul46.3}$\pm$0.4            & 42.8$\pm$1.3             & 39.3$\pm$0.6      & \textbf{71.0}$\pm$0.8   \ (\ $\uparrow$ 53.3\%)  \\
\bottomrule
\end{tabular}
}
% \vspace{-3pt}
\end{table*}

% \usepackage{multirow}

\begin{figure}[t]
\centering
\includegraphics[width=0.65\linewidth]{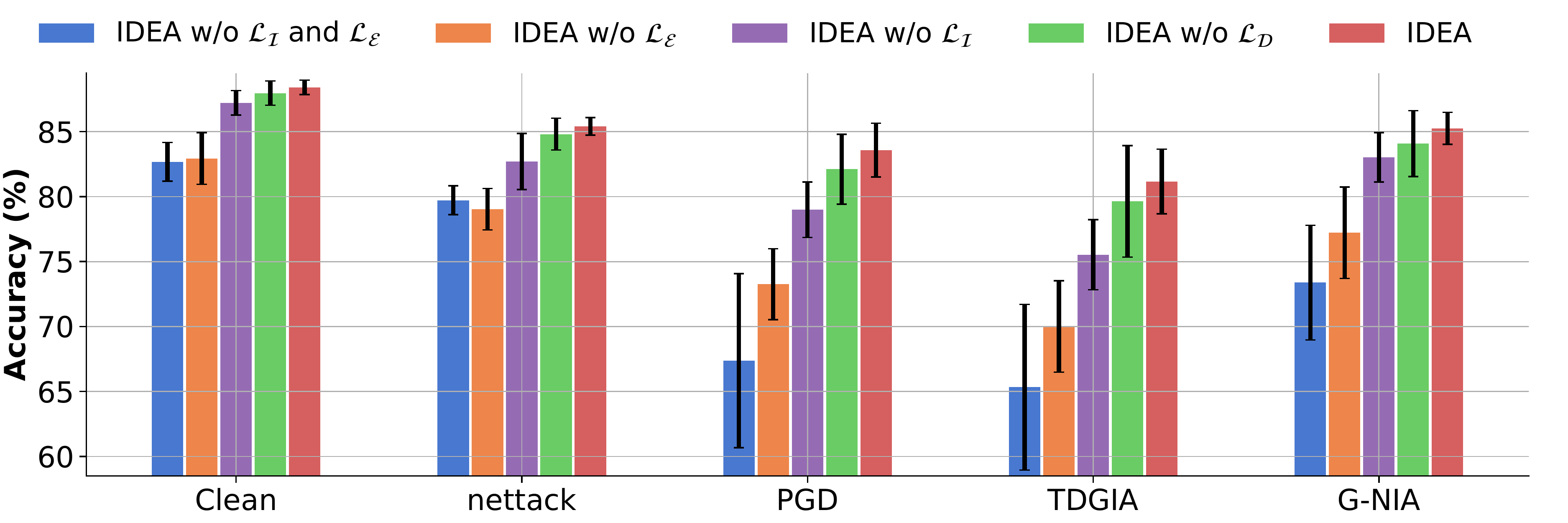}
\caption{Ablation Study.}
\label{fig:abl}
\end{figure}

\subsection{Robustness against Poisoning Attacks}
We evaluate IDEA's robustness under poisoning attacks, employing the widely-adopted MetaAttack~\cite{zugner_adversarial_2019} and varying the perturbation rate (the rate of changing edges) from 0 to 20\% following~\cite{liu2021elastic,li2022reliable}. We exclude ogbn-arxiv since MetaAttack cannot handle large graphs.
Table~\ref{tab:poison} shows that all methods' accuracy decreases as perturbation rate increases. Among baselines, graph purification methods demonstrate better defense performance, with STABLE outperforming others on Cora and Citeseer. RGCN, SimPGCN, and Soft-Median resist attacks only at low perturbation rates. The adversarial training FLAG brings less improvement than it does under  evasion attacks. Maybe due to its training on evasion attacks, leading to poor generalization for poisoning attacks.

IDEA achieves the state-of-the-art performance under all perturbation rates on all datasets, significantly outperforming all baselines. 
%IDEA has a huge improvement compared with the second best-performing method.
When the attack strength becomes larger, our IDEA still maintains good performance, demonstrating that IDEA has the invariant prediction ability across perturbations.

%The results on Table~\ref{tab:evasion} and Table~\ref{tab:poison} both demonstrate that IDEA achieves adversarial robustness against both various evasion attacks and poisoning attacks with multiple perturbation rates.

%\begin{figure}[t]
%%\centering
%\begin{minipage}[t]{0.35\textwidth}
%%\centering
%\includegraphics[width=1\linewidth]{ablation.pdf}
%\caption{Ablation Study.}
%\label{fig:abl}
%\end{minipage}
%\hfill
%\begin{minipage}[t]{0.58\textwidth}
%%\centering
%\includegraphics[width=1\linewidth]{visual.jpg}
%\caption{{Visualizing learned features: clean and attacked graphs.}
%\label{fig:vis}
%\end{minipage}
%\end{figure}

%\begin{figure}[t]
%\centering
%\begin{minipage}[t]{0.48\textwidth} % 调整左侧 minipage 的宽度
%    \centering
%    \includegraphics[width=\linewidth]{ablation.pdf} % 使图片填满 minipage 宽度
%    \captionsetup{justification=raggedright} % 确保图题在 minipage 中居中
%    \caption{Ablation Study.}
%    \label{fig:abl}
%\end{minipage}
%\hfill
%\begin{minipage}[t]{0.48\textwidth} % 右侧 minipage 保持不变
%    \centering
%    \includegraphics[width=\linewidth]{visual.jpg}
%    \captionsetup{justification=centering} % 同样确保图题在 minipage 中居中
%    \caption{\small{Visualizing learned features: clean and attacked graphs.}}
%    \label{fig:vis}
%\end{minipage}
%\end{figure}

\subsection{Ablation Study}
\label{sec:abl}
We analyze the influence of each part of IDEA through experiments on invariance goals and domain construction. 
We implement four variants of IDEA, including IDEA without node-based invariance goal (IDEA w/o $\mathcal{L}_\mathcal{I}$), IDEA without structure-based invariance goal $\mathcal{L}_\mathcal{E}$ (IDEA w/o $\mathcal{L}_\mathcal{E}$), IDEA without both invariance goals (IDEA w/o $\mathcal{L}_\mathcal{I}$ and $\mathcal{L}_\mathcal{E}$), and IDEA without domain partition (IDEA w/o $\mathcal{L}_\mathcal{D}$).
The variant IDEA w/o $\mathcal{L}_\mathcal{I}$ and $\mathcal{L}_\mathcal{E}$, which only optimizes the predictive loss, serves as a test for the benefit brought by learning causal features.
We  take the results on clean graph and evasion attacks on Cora as an illustration.

As shown in Figure~\ref{fig:abl}, all variants exhibit a decline compared to IDEA (red), highlighting the significance of both invariance goals and domain construction. 
Specifically, IDEA w/o $\mathcal{L}_\mathcal{I}$ and $\mathcal{L}_\mathcal{E}$ (blue) suffers the largest drop, highlighting our objectives' benefits since IDEA is much more robust than simple adversarial training using same adversarial examples. 
The performance decline of IDEA w/o $\mathcal{L}_\mathcal{E}$ (orange) illustrates the significant advantages of the structure-based invariance goal, especially on clean graph, highlighting the benefits of modeling the interactions between samples.
% This may be due to the first term in $\mathcal{L}_\mathcal{E}$ aiming to predict node $i$’s label correctly based on $k$'s representation, 
% implying local smoothness assumption and thus improving performance on clean graphs.
IDEA w/o $\mathcal{L}_\mathcal{D}$ (green) displays a large standard deviation, with the error bar much larger than that of IDEA, emphasizing the stability achieved through the diverse attack domains.

%The ablation study demonstrates the effectiveness of both invariance objectives and domain construction of IDEA.
%Hyper-parameter analysis regarding $\alpha$ and the number of attack domains is in Appendix D.5,
%~\ref{apd:para}. 
%while Appendix D.6 details the performance under adaptive attacks .
%~\ref{apd:adaptive}.

%\begin{wrapfigure}{r}{0.53\textwidth}

\begin{figure}[t]
   \captionsetup[subfigure]{font=scriptsize,labelfont=scriptsize}
   \begin{subfigure}{.42\textwidth}
       \includegraphics[width=1\linewidth]{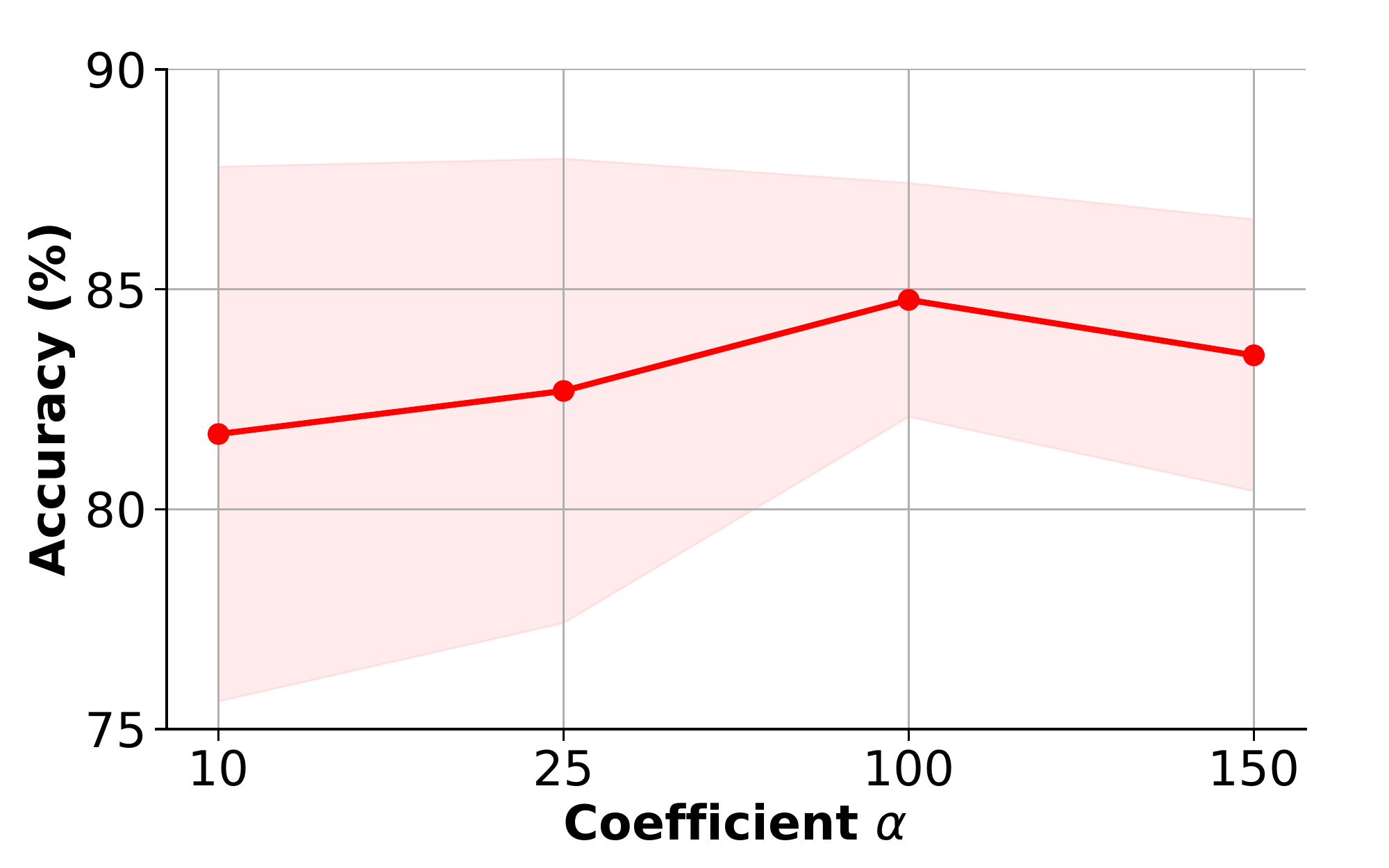}
       % \caption{Hyperparameter analysis for the coefficient $\alpha$.}
       \label{subfig:hyper_alpha}
   \end{subfigure}
   \hspace*{\fill}
   \begin{subfigure}{.42\textwidth}
       \includegraphics[width=1\linewidth]{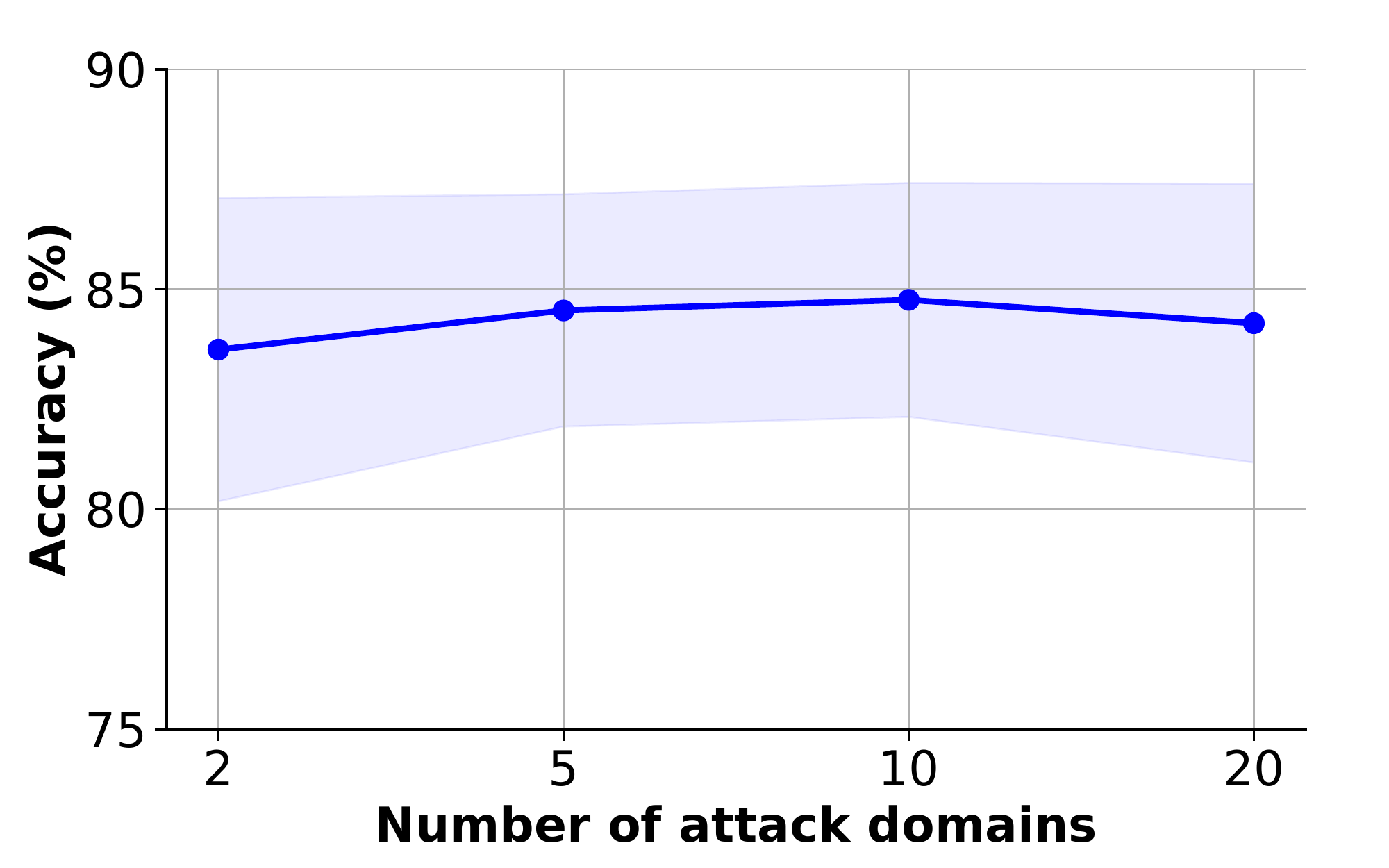}
       % \caption{Hyperparameter analysis for the number of attack domains.}
       \label{fig:hyper_env}
   \end{subfigure} 
   \caption{Hyperparameter analysis: The average accuracy of clean and attacked graphs, including Clean, nettack, PGD, TDGIA, and G-NIA.}
   \label{fig:hyper}
   \vspace{10pt}
\end{figure}

\subsection{Hyper-Parameter Analysis}
\label{apd:para}

We investigate the effects of coefficient $\alpha$ and the number of domains and compare the defense performance. Note that we take results against evasion attacks on Cora as an illustration. Figure~\ref{fig:hyper} shows that the average accuarcy of the clean and attacked graphs, along with the standard deviation of accuracy across these graphs, i.e., AVG in Section 4.1. For coefficient $\alpha$, we observe that when $\alpha$ is increasing, IDEA achieves better performance (higher accuracy), and performs more stable (lower standard deviation), validating the effectiveness of invariance component. While, too large $\alpha$ (e.g. $\alpha=150$) causes domination of invariance goal, leading to little attention to the predictive goal and degradation of performance. Regarding the number of attack domains, performance improves with increasing domain numbers, reaching its peak at 10 domains. This may be due to the relatively small number of nodes in the Cora dataset, suggesting that a larger number of domains (e.g., 20) is not necessary. In our main experiment shown in Table 1, we utilized $\alpha=100$ and the attack domain number to 10 to achieve the best results.

\begin{table}[t]
   \centering
   \caption{Accuracy(\%) of targets under adaptive attack.}
   \label{tab:adaptive}
   \resizebox{0.7\textwidth}{!}{
   \begin{tabular}{cccccccc}
   \toprule
       \textbf{Dataset} & \textbf{GCN} & \textbf{ProGNN} & \textbf{STABLE} & \textbf{RGCN} & \textbf{SimPGCN} & \textbf{FLAG} & \textbf{IDEA} \\ 
       \midrule
       Cora & 18.7$\pm$3.5 & 15.0$\pm$2.8 & 27.5$\pm$5.0 & 14.3$\pm$1.6 & 28.9$\pm$3.4 & 36.7$\pm$2.4 & \textbf{53.1}$\pm$5.0 \\ 
       Citeseer & 11.8$\pm$2.0 & 21.8$\pm$2.3 & 12.7$\pm$2.3 & 10.1$\pm$1.2 & 24.5$\pm$4.7 & 31.1$\pm$6.4 & \textbf{44.4}$\pm$1.6 \\ 
       Reddit & 34.7$\pm$4.9 & 43.1$\pm$8.1 & 27.3$\pm$4.4 & 57.5$\pm$3.0 & 12.5$\pm$6.7 & 5.2$\pm$5.9 & \textbf{61.7}$\pm$5.3 \\ 
   \bottomrule
   \end{tabular}
   }
   \label{tab:adaptive}
   \vspace{10pt}
\end{table}

\begin{figure}[t]
\centering
\includegraphics[width=0.65\linewidth]{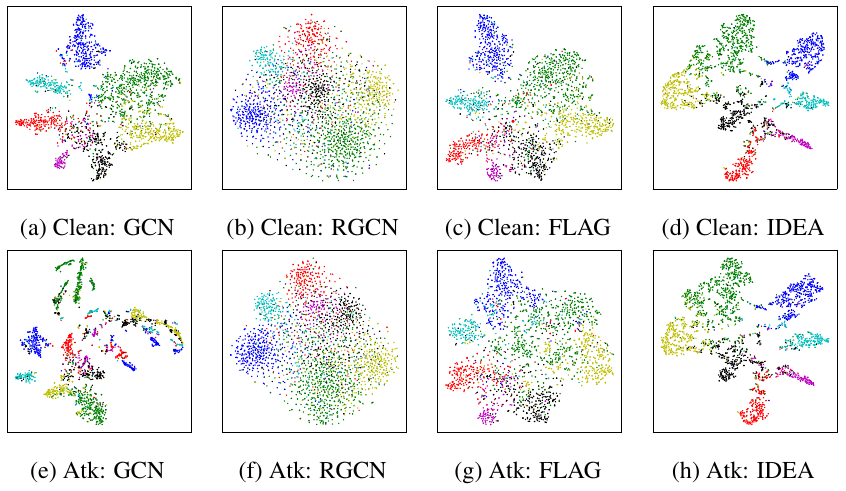}
\caption{Visualizing learned features: clean and attacked graphs. 
Color denote ground-truth labels.
}
\label{fig:vis}
\end{figure}
\subsection{Performance under Adaptive Attacks}
\label{apd:adaptive}
To better evaluate our IDEA, we also conduct experiments under adaptive attack, i.e., PGD in~\cite{aredefenserobust}.
We implement adaptive attacks for baselines and IDEA. 
Some baselines are excluded because their open source codes use edge\_index to represent edges. This makes calculating gradients on edges challenging, so the implementation of these baselines are difficult to conduct white-box adaptive attacks. 
As shown in Table~\ref{tab:adaptive}, adaptive attack causes serious performance degradation to defense methods because adaptive attacks are powerful white-box attacks. 
IDEA outperforms all the baselines. Experiments offer a more broader evaluation of IDEA's performance under a hard scenario, consistently showing IDEA's superiority.

\subsection{Visualization}
We further visualize the learned features with t-SNE technique~\cite{van2008visualizing} to show whether IDEA learns the features that have strong and invariant predictability. Figure~\ref{fig:vis} illustrates the feature learned by GCN, RGCN, FLAG, and IDEA on clean graph and under the strongest G-NIA attack on Cora. As shown in Figure~\ref{fig:vis}, existing methods either learn the features destroyed under attack (GCN  and FLAG), or learn features are mixed (RGCN).

% % For IDEA, we adopt the representation $h(G)_i$ in Eq.~\ref{eq:final_loss}. For the methods that do not have separate predictors, we visualize the learned representation of the second to last layer.

% %(a,e),  GCN has learned discriminative features for labels on clean graph, but under attack, the learned feature seems confusing and miscellaneous.
% %Figure~\ref{fig:vis}(b,f) shows that the STABLE learns discriminative representations on Clean, however, the representations are seriously destroyed under attack, so they are no longer discriminative to predict labels.
% %RGCN (Figure~\ref{fig:vis}(b,f)) and FLAG (Figure~\ref{fig:vis}(c,g)) learn features of different labels mix to some extent, resulting in the decline of performance.

For IDEA, in Figure~\ref{fig:vis}(d,h), the features learned by IDEA can be distinguished by labels. Specifically, IDEA's learned features are similar for nodes with the same label and distinct for different labels, emphasizing features' \textbf{strong predictability} for labels. Furthermore, the features in Figure~\ref{fig:vis}(d) on clean graph and those in Figure~\ref{fig:vis}(h) on attacked graph exhibit nearly the same distributions. This observation demonstrates that the relationship between features and labels remained invariant across attacks, thus exhibiting \textbf{invariant predictability}. 
These results show that IDEA learned causal invariant features with both strong and invariant predictability for labels.

\section{Related works}
\label{apd:related}
In this section, we present the related works on defense methods against graph adversarial attacks and invariant learning methods.

\subsection{Defense against Graph Adversarial Attack}
Despite the success of graph neural networks (GNNs), they are shown to be vulnerable to adversarial attacks,~\cite{zugner2018adversarial,Sun2018AdversarialAA,Chen2020ASO,gaorumor,gaoalleviating,gaoaddressing},  i.e., imperceptible perturbations on graph data can  dramatically degrade the performance of GNNs~\cite{TaoGNIA,ZouTDGIA,Sun2020AdversarialAO,Wang2020ScalableAO,Tao2023CANA}, blocking the deployment of GNNs to real world applications~\cite{Jin2020AdversarialAA}. Various defense mechanisms~\cite{li2023revisiting,gosch2023revisiting,Tao2023AdvImmune,Tao2021advimmune} have been proposed to counter these graph adversarial attacks, which can broadly be classified into adversarial training, graph purification, and robust aggregation strategies~\cite{Sun2018AdversarialAA,li2022reliable,Jin2020AdversarialAA}. 

% Adversarial training methods, such as AdvT~\cite{Dai2019AdversarialTM}, FLAG~\cite{kong2020flag}, and others~\cite{feng2019graph,  li2022spectral}, generally adopt a min-max optimization style, iteratively generating adversarial examples that maximize the loss and updating GNNs parameters that minimize the loss on the seen adversarial examples. However, the pattern varies with different attacks, leading to the non-robustness of adversarial training to unseen attacks~\cite{Bojchevski2019CertifiableRT}.  Later, robust training~\cite{Zgner2019CertifiableRA,Bojchevski2019CertifiableRT,zugner2020CertiRob,bojchevski_sparsesmoothing_2020, scholten2022randomized} is proposed to further include the worst-case adversarial examples and improve certifiable robustness~\cite{Bojchevski2019CertifiableRT, Zgner2019CertifiableRA}, which can be seen an improved version of traditional adversarial training. However, due to the limited search space, roust training still suffers the same problem as adversarial training. 

Adversarial training methods, such as FLAG~\cite{kong2020flag} and others~\cite{Dai2019AdversarialTM,feng2019graph,li2022spectral}, typically employ a min-max optimization approach. This involves iteratively generating adversarial examples that maximize the loss and updating GNN parameters to minimize the loss on these examples. However, adversarial training may be not robust under unseen attacks~\cite{Bojchevski2019CertifiableRT}. 
% To address this issue, 
Robust training methods~\cite{Zgner2019CertifiableRA,Bojchevski2019CertifiableRT,zugner2020CertiRob,bojchevski_sparsesmoothing_2020,scholten2022randomized}  incorporate worst-case adversarial examples to enhance certifiable robustness~\cite{Bojchevski2019CertifiableRT,Zgner2019CertifiableRA}. These methods can be considered an improved version of traditional adversarial training. However, due to limited searching space, robust training still faces similar challenges as adversarial training.

Graph purification methods~\cite{Wu2019AdversarialEF,Jin2020GraphSL,Entezari2020AllYN} aim to purify adversarial perturbations by modifying graph structure. 
Jaccard~\cite{Wu2019AdversarialEF} prunes edges that connect two dissimilar nodes, while ProGNN~\cite{Jin2020GraphSL} concurrently learns the graph structure and GNN parameters through optimization of  feature smoothness, low-rank and sparsity. 
The recent method STABLE~\cite{li2022reliable} acquires reliable representations of graph structure via unsupervised learning.
GARNET~\cite{GARNET22DengLF} first leverages weighted spectral embedding to construct a base graph, then refines the base graph by pruning additional uncritical edges based on probabilistic graphical model, to boost the adversarial robustness of GNN models.

Robust aggregation methods~\cite{Zhu2019RobustGC,liu2021elastic,jin2021node,lei2022evennet,ZhangGNNGuard2020}  redesign model structures  to establish robust GNNs. 
RGCN~\cite{Zhu2019RobustGC} uses Gaussian noise to mitigate adversarial perturbations. 
% GNNGuard~\cite{ZhangGNNGuard2020} quantifies and assigns weight to edges.
SimPGCN~\cite{jin2021node} resents a feature similarity preserving aggregation that balances the structure and feature information.
Elastic~\cite{liu2021elastic} improves the local smoothness adaptivity and derives the elastic message passing.
Geisler et al.~\cite{GeislerSSZBG21} design a robust aggregation function, Soft Median to achieve an effective defense at all scales. 
However, both kinds of methods rely on specific heuristic priors such as local smoothness~\cite{Wu2019AdversarialEF,velickovic2018graph,Jin2020GraphSL,ZhangGNNGuard2020,li2022reliable,jin2021node} or low rank~\cite{Jin2020GraphSL,Entezari2020AllYN}, that may be ineffective against some attacks~\cite{chen2022understanding}, leading to method failure. 
What's worse, modifying graph structure~\cite{Jin2020GraphSL,ZhangGNNGuard2020} or adding noise~\cite{Zhu2019RobustGC} with this heuristic may even cause performance degradation on clean graphs. 

Different from the above studies, in this paper, we creatively propose an invariant causal defense perspective, providing a new perspective to address this issue. Our method aims to learn causal features that possess strong predictability for labels and invariant predictability across attacks, to achieve graph adversarial robustness.

% \subsubsection{Adversarial training methods}
% \label{sec:at}
% The adversarial training methods have been widely used to defending against adversarial attack and improve adversarial robustness.

% It is more like a interpolations of the seen adversarial examples~~\cite{arjovsky2019invariant}.
% For eror on training data

% Our can achieve the extrapolation to all 

% \subsubsection{Robustness certification}
% The

\subsection{Invariant Learning Methods}
Invariant learning methods~\cite{arjovsky2019invariant,krueger2021out,li2022invariant} have fueled a surge of research interests~\cite{shen2021towards,creager2021environment,yong2022zin,chen2022does,chenlearning,wu2022handling}. 
These work typically assume that data are collected through different domains or environments~\cite{arjovsky2019invariant}, and the causal relationships within the data remain unchanged across different domains, denoting invariant causality~\cite{shen2021towards}. 
Generally, invariant learning methods aim to learn the causal mechanism or causal feature that is invariant across different domains or environments, allowing the causal feature to generalize across all domains, which can be used to solve the out-of-distribution generalization problem~\cite{shen2021towards,arjovsky2019invariant,rosenfeld2020risks}. 

However, such methods cannot be directly applied to solve graph adversarial robustness due to the complex nature of graph data and the scarcity of diverse domains. Two main challenges arise:
% The complex nature of graph data and the lack of diverse domains bring great challenges.
i) On graph data, there are interconnections (edges) between nodes, so nodes are no longer independent of each other, making samples not independent and identically distributed (non-IID)~\cite{wu2022handling,chenlearning}. We model the generation of graph adversarial attack via an interaction causal model and propose corresponding invariance goals considering both node itself and the interconnection between nodes. 
ii) In adversarial learning, constructing sufficiently diverse domains or environments is challenging due to a lack of varied domains. We propose to learn sufficient and diverse domains by limiting the co-linearity between domains.

\subsection{Causal methods for Adversarial Robustness}
A few recent works attempt to achieve adversarial robustness with causal methods  on computer vision~\cite{ren2022dice,zhangadversarial}.
These methods, such as DICE~\cite{ren2022dice}, ADA~\cite{zhangadversarial}, mainly use causal intervention to achieve the robustness.
The difference between them and our work lies in two aspects:
(1) Existing causal methods for robustness are developed for the image area. However, the non-IID nature of graph data brings challenges to these methods in achieving graph adversarial robustness. Our work proposes the structural-level invariance goal for the non-IID graph data.
(2) These methods adopt causal intervention. For example, DICE uses hard intervention~\cite{ren2022dice}, and ADA~\cite{zhangadversarial} uses "soft" intervention.
However, the intervention is difficult to achieve~\cite{pearl2009causal}. Our work constructs diverse domains and learns causal features by optimizing both node-based and structural-based invariance goals.

\subsection{Purification methods in Computer Vision }
{
There are also some purification works in computer vision for defending against attacks.
Shi et al. propose Self-supervised Online Adversarial Purification (SOAP), leveraging self-supervised loss to purify adversarial examples at test-time~\cite{soap}.
Zhou et al. propose to remove adversarial noise by implementing a self-supervised adversarial training mechanism in a class activation feature space~\cite{CAFD}.
Naseer et al. propose a self-supervised adversarial training mechanism in the input space~\cite{NRP}.
Liao et al. propose high-level representation guided denoiser (HGD), using a loss function defined as the difference between the target model’s outputs activated by the clean image and denoised image~\cite{HGD}.
}

{
Most purification methods in computer vision leverage image data priors. For example, SOAP~\cite{soap} incorporates self-supervised tasks such as image rotation that are unique to the domain of computer vision, while NRP~\cite{NRP} depends on a pixel loss function, i.e., $\mathcal{L}_{img}$, to encourage image smoothness. These domain-specific dependencies pose significant challenges when considering the direct transposition of these methods to graph data, which inherently lacks such image-based priors.}

{
In contrast, graph purification methods~\cite{Jin2020GraphSL,li2022reliable} are specifically designed to exploit the unique properties of graph data, making them appropriate for addressing graph-specific issues. However, graph purification defenses rely on predefined heuristics, while these may be ineffective for som attacks causing the methods to fail.
Therefore, there is a pressing need to develop a defense strategy that is robust and effective against various attacks.
}

% % \subsection{Discussion on Graph Adversarial Robustness and OOD Generalization}
% % The graph adversarial robustness and OOD generalization are both minmax problem. The

%% \vspace{-2pt}
\section{Conclusion and Future work}
% % \section{Limitation and Broader impact}

% % There are several valuable research questions, such as designing better domain partitioning, or interpreting the causal feature
% % studying theoretical guarantee for general scenarios.

% % our method may encounter difficulties when applied to very large-scale graphs (tens of millions of nodes) due to the sampling neighbor operation in structure-based invariance goal, which can increase memory consumption for node with numerous neighbors. Still, our approach can handle large datasets with over 100,000 nodes and million-scale edges, such as ogbn-arxiv, demonstrating  positive outcomes. Future work will focus on investigating more efficient structural attack strategies for generating adversarial samples.

% Moreover, interpreting graph data proves challenging, making it difficult to identify causal features analogous to those in image data. Our study aims at visualizing the relationships between causal features and labels, striving to showcase that these causal variables possess both strong predictability for labels and invariant predictability across attacks.

In this paper, we creatively introduce a causal defense perspective by learning causal features that have strong and invariant predictability across attacks.
%We propose an invariant causal defense method IDEA, by learning the causal feature with strong and invariant predictability across attacks to achieve graph adversarial robustness.
Then, we propose IDEA and design two invariance objectives to learn causal features. 
%To characterize the non-IID graph data, we model the generation of graph adversarial attack via an interaction causal model and propose node-based and structure-based invariance objectives accordingly. 
%To construct domains, we propose to learn sufficient and diverse domains by limiting the co-linearity between domains. 
Extensive experiments demonstrate that IDEA significantly outperforms all the baselines under both evasion attacks and poisoning attacks on five benchmark datasets, emphasizing that IDEA possesses both strong and invariant predictability across attacks.
%The ablation studies demonstrate the effectiveness of our proposed two invariance losses and the domain construction component.
% We believe the causal defense perspective is a promising research direction,  
% and look forward to further exploration in the future.
%
% \textbf{Broader impact}. Our paper focuses on a defense method, aiming to resist adversarial attacks and enhance the reliability of Graph Neural Networks (GNN) in practical applications. We do not foresee any significant negative impact associated with this research.
% \textbf{Limitation}. Our work mainly focuses on graph data. It is challenging for us to identify causal features like those in image data,  due to its inherent complexity and visualization limitations. Instead, we visualize the relationships between causal features and labels, aiming to show that these causal variables can have both strong predictability for labels and invariant predictability across attacks. In addition, although empirical evidence supports IDEA's superiority across various datasets, theoretically, we can only establish invariant defender when assuming a linear causality. We believe studying theoretical guarantee for general scenarios is a valuable research question.
%To the best of our knowledge, this paper is the first work to solve the graph adversarial attack and defense problem from the perspective of invariant causal learning and is expected to bring new insights into the graph adversarial learning area. 
We believe causal defense approach is a promising new direction, and there are many interesting and valuable research problems in the future. For example, studying domain partitioning is more suitable for adversarial attack and defense scenarios; or exploring more ways to generate adversarial examples.

\section*{Acknowledgments}

This work is funded by the National Key R\&D Program of China (2022YFB3103700, 2022YFB3103701), the Strategic Priority Research Program of the Chinese Academy of Sciences under Grant No. XDB0680101, and the National Natural Science Foundation of China under Grant Nos. 62102402, U21B2046, 62272125. Huawei Shen is also supported by Beijing Academy of Artificial Intelligence (BAAI).


%\bibliographystyle{model1a-num-names}
\bibliographystyle{cas-model2-names}
%\bibliographystyle{elsarticle-num}

%\bibliography{IDEA.bib}
%\input{IDEA.bbl}

\clearpage
\appendix
%%
%%
%\subsection{Attack Performance on Vanilla GNN}
%\label{apd:vanilla}
%Figure~\ref{tab:vanilla} shows  that the attack performance on vanilla GCN without any defense or detection methods on three datasets.
%
%\begin{table}[]
%\caption{Misclassification rate (\%) of node injection attacks on GCN. Larger is better, and the largest one is bolded}
%\label{tab:vanilla}
%\begin{tabular}{c|c|c|c}
%\toprule
%      & ogbn-products & reddit             & ogbn-arxiv         \\
%\midrule
%Clean & 21.01                         & 8.15            & 28.49          \\
%\midrule
%G-NIA  & \textbf{99.95}                & 99.85           & \textbf{98.81} \\
%+HAO  & 99.81                         & \textbf{100.00} & 98.48          \\
%+CANA & 76.66                         & 63.82           & 45.48          \\
%\midrule
%TDGIA & 93.95                         & \textbf{99.80}  & 79.36          \\
%+HAO  & \textbf{97.00}                & 97.25           & \textbf{80.57} \\
%+CANA & 52.74                         & 42.63           & 49.02          \\
%\midrule
%PGD   & \textbf{97.52}                & \textbf{99.80}  & \textbf{99.93} \\
%+HAO  & 91.38                         & 94.80           & 93.84          \\
%+CANA & 90.52                         & 57.92           & 53.12   \\
%\bottomrule      
%\end{tabular}
%\end{table}

% \subsection{Performance on clean graphs}

\section{{Symbol and Definition}}
This section summarizes all symbols and their definitions for a clear understanding of our paper.
Understanding the notations and terminologies used throughout this paper is crucial for comprehending the theoretical constructs and the methodologies proposed. To facilitate this, Table~\ref{tab:symbols} shows a comprehensive summary of all the symbols and their associated definitions, categorized into three main parts for ease of reference.
%Table~\ref{tab:symbols} summarizes all symbols and their definitions for quick reference. 

Specifically, the first part of the table presents the foundation symbols related to graph, including nodes and edges. 
The second part lists the symbol regarding interaction causal model in Section~\ref{sec:causal}.
In the third part, we delve into the specific notation employed in the proposed IDEA, including components, representation and loss functions in our IDEA method.

\begin{table}
\centering
\caption{{Symbol table}}
\label{tab:symbols}
%\rowcolors{2}{lightblue}{white} % Alternating row colors starting from the 2nd row
\begin{tabular}{c p{10cm}}
\toprule
\textbf{Symbol} & \textbf{Definition} \\
\midrule
$G$ & Graph in a node classification task \\
$\mathcal{V}$ & Node set of a graph \\
$\mathcal{E}$ & Edge set of a graph \\
$X$ & Attribute matrix \\
$\mathcal{K}$ & Class set \\
$K$ & Class number \\
$f_{\theta}$ & GNN classifier \\
$\hat{G}$ & Perturbed graph \\
$\mathcal{G}$ & Admissible perturbed graph set \\
$i$, $j$ & Nodes in graph \\
\midrule
$G_i$ & Input ego-network of node $i$ \\
$Y_i$ & Label of node $i$ \\
$C_i$ & Causal feature of node $i$ \\
$D_i$ & Attack domain of node $i$ \\
$N_i$ & Non-causal feature of node $i$ \\
\midrule
$I(\cdot)$ & Mutual information \\
$\Phi$ & Feature encoder \\
$C_\mathcal{N}$ & Causal feature of neighbor $\mathcal{N}$ \\
$Z$ & Representation of feature encoder \\
$p(\cdot)$ & Natural distribution \\
$q(\cdot)$ & Variation approximation \\
$h$ & Neural network feature encoder \\
$g$ & Neural network classifier \\
$g_d$ & Neural network auxiliary classifier \\
$s$ & Domain learner \\
$\mathcal{V}^{D}$ & Nodes assigned to domain $D$ \\
$r^{D}$ & Overall representation of $\mathcal{V}^{D}$ \\
$\mathcal{L}_\mathcal{P}$ & Predictive loss \\
$\mathcal{L}_\mathcal{I}$ & Node-based invariance loss \\
$\mathcal{L}_\mathcal{E}$ & Structure-based invariance loss \\
$\mathcal{L}_\mathcal{D}$ & Domain loss \\
$\gamma$ & Intrinsic causal mechanism \\
$\epsilon$ & Gaussian noise \\
$\psi$ & Mapping from causal and non-causal features to graph representation \\
$\rho$ & Powerful graph representation extractor \\
$\Theta_{\cdot}$ & Parameter associated with a particular model component \\
\bottomrule
\end{tabular}
\end{table}

\section{Proofs}
\subsection{Proof for Proposition 1}

\begin{proof}
\label{proof:cond_info}
	The difference between $\hat{I}(Y,D|Z)$ and $I(Y,D|Z)$ could be written as
\begin{equation}
\begin{aligned}
&\hat{I}(Y,D|Z)-I(Y,D|Z)\\ 
=&\ 
\mathbb{E}_{p(z)}\left[ \mathbb{E}_{p\left(y,d|Z\right)}\left[\left[\log q_d(y \mid z, d) - \log q(y|z)\right] - \left[\log p(y \mid z, d)-\log p(y \mid z)\right] \right] \right]\\
=&\ \mathbb{E}_{p(z)}\left[ \mathbb{E}_{p(y,d \mid z)} \left[ \log \frac{p(y \mid z)}{q(y \mid z)}- \log \frac{p(y \mid z, d)}{q_d(y \mid z, d)}\right] \right]\\
=&\ \mathbb{E}_{p(z)}\left[ \mathbb{E}_{p(y \mid z)} \left[ \log \frac{p(y \mid z)}{q(y \mid z)} \right]- \mathbb{E}_{p(d \mid z)} \mathbb{E}_{p(y \mid z,d)} \left[ \log \frac{p(y \mid z, d)}{q_d(y \mid z, d)}\right] \right]\\
=&\ \mathbb{E}_{p(z)} KL\left[p(y \mid z) \| q(y\mid z)\right]- \mathbb{E}_{p(z,d)}
KL\left[p(y \mid z, d) \| q_{d}(y \mid z, d)\right]
\end{aligned}
\end{equation}

Next, similar to the theoretical analysis in CLUB~\cite{club2020}, we can prove that $\hat{I}$ is either a upper bound of $I$ or a esitimator of $I$ whose absolute error is bounded by the approximation performance $\mathbb{E}_{p(z,d)} KL[p(y \mid z, d) \| q_{d}(y \mid z, d)]$. That is to say, if $\mathbb{E}_{p(z,d)} KL[p(y \mid z, d) \| q_{d}(y \mid z, d)]$ is small enough, $I(Y,D|Z)$ is bounded by $\hat{I}(Y,D|Z)$. 
Therefore, $\hat{I}(Y,D|Z)$ is minimized if $\mathbb{E}_{p(z,d)} KL[p(y \mid z, d) \| q_{d}(y \mid z, d)]$ and $\hat{I}(Y,D|Z)$ are both minimized.
\end{proof}

\subsection{Proof for Proposition 2}
\label{proof:achiev}
\begin{proof}
The proof includes three steps: 

\textbf{Step 1:} We prove that if $\Phi$ and $\omega$ satisfies the condition (1), 
i.e., {\small$I\left(\Phi(\hat{G}), Y\right)- \left[I\left(Y, D \mid \Phi(\hat{G})\right)+ I\left(Y, D \mid \Phi(\hat{G})_{\mathcal{N}}\right)\right]$} (denoted as $\kappa$) is maximized, 
then {\small$\Theta_{\phi} \mathbb{E}_{\rho(\hat{G})^D}\left[\rho(\hat{G}^{D}) \rho(\hat{G}^{D})^{ \top}\right]
\Theta_{\phi}^\top \Theta_\omega=\Theta_{\phi} \mathbb{E}_{\rho(\hat{G}^{D}),{Y}^{D}}\left[\rho(\hat{G}^{D}){Y}^{D}\right]$} 
where {\small $\hat{G}^{D}=\left\{\rho(\hat{G})_i|i\in \mathcal{V}^D\right\}$}, for all $D\in \mathcal{D}_\text{tr}$. 
%Suppose that $\Phi$ has infinite capacity for representation, with $\Phi= \arg\max \kappa$, we have $Y^D=\Phi(\hat{G})\cdot \gamma + \epsilon_{\Phi}$, where the error term $\epsilon_{\Phi}$ satisfies $\mathbb{E}_{{Y}^{D}}\left[\epsilon_{\Phi}\right]=0$. 
%Note that we denote $\Phi(\hat{G})$ in $\kappa$ as 
%$\Phi(\hat{G}) \triangleq {\hat{G}^{{D}^\top}}\Theta_{\phi}^\top \Theta_\omega$.
Next, we begin our proof.
Suppose that $\Phi$ has infinite capacity for representation, with $\Phi= \arg\max \kappa$ and $\Phi$ containing $\phi$ and $\rho$, 
we have $Y^D={\rho(\hat{G}^{D})^{\top}} \Theta_{\phi}^\top \Theta_\omega + \epsilon_{\Phi}$, where ${\rho(\hat{G}^{D})^{\top}}\Theta_{\phi}^\top \Theta_\omega$ represents the output of $\rho(\hat{G}^{D})$ after passing through learner $\phi$ and classifier $\omega$ (i.e., the output of $\hat{G}^{D}$ after passing through $\Phi$ and $\omega$).
The error term $\epsilon_{\Phi}$ satisfies $\mathbb{E}_{{Y}^{D}}\left[\epsilon_{\Phi}\right]=0$. 
We have:
\begin{equation}
%\small
\begin{aligned}
\label{eq:p2_s1}
   Y^D &= {\rho(\hat{G}^{D})^{\top}}\Theta_{\phi}^\top \Theta_\omega + \epsilon_{\Phi}  \\
   \mathbb{E}_{{Y}^{D}}\left[Y^D\right] &= 
   \mathbb{E}_{Y^{D}}\left[{\rho(\hat{G}^{D})^{\top}}\Theta_{\phi}^\top \Theta_\omega + \epsilon_{\Phi}\right] = \mathbb{E}_{Y^{D}}\left[{\rho(\hat{G}^{D})^{\top}}\Theta_{\phi}^\top \Theta_\omega\right] = {\rho(\hat{G}^{D})^{\top}}\Theta_{\phi}^\top \Theta_\omega\\
   \mathbb{E}_{\rho(\hat{G}^{D}),{Y}^{D}}\left[\rho(\hat{G}^{D}){Y}^{D}\right]&=  \mathbb{E}_{\rho(\hat{G}^{D})}\left[\rho(\hat{G}^{D}){\rho(\hat{G}^{D})^{\top}}\right]\Theta_{\phi}^\top \Theta_\omega\\
   \Theta_{\phi} \mathbb{E}_{\rho(\hat{G}^{D}),{Y}^{D}}\left[\rho(\hat{G}^{D}){Y}^{D}\right]&= 
   \Theta_{\phi} \mathbb{E}_{\rho(\hat{G}^{D})}\left[\rho(\hat{G}^{D}){\rho(\hat{G}^{D})^{\top}}\right]\Theta_{\phi}^\top \Theta_\omega.\\
\end{aligned}
\end{equation}

The validity of line 2 in Eq.~\ref{eq:p2_s1} stems from $\mathbb{E}_{{Y}^{D}}\left[\epsilon_{\Phi}\right] = 0$, and the fact that $Y^D$ is independent with ${\rho(\hat{G}^{D})^{\top}}\Theta_{\phi}^\top \Theta_\omega$. Consequently, we have $\Theta_{\phi} \mathbb{E}_{\rho(\hat{G}^{D})}\left[\rho(\hat{G}^{D}){\rho(\hat{G}^{D})^{\top}}\right]\Theta_{\phi}^\top \Theta_\omega = \Theta_{\phi} \mathbb{E}_{\rho(\hat{G}^{D}),{Y}^{D}}\left[\rho(\hat{G}^{D}){Y}^{D}\right]$.

\textbf{Step 2:} We prove that if $\Theta_{\phi}^\top \Theta_\omega$ satisfies the condition (2), 
i.e.,  {\small$\left\{\mathbb{E}_{\rho(\hat{G}^{D})}\left[\rho(\hat{G}^{D}) \rho(\hat{G})^{D ^\top}\right]
\left(\Theta_{\phi}^\top \Theta_\omega - \Theta_{\tilde{\psi}}^\top \Theta_\gamma\right)\right\}_{D\in \mathcal{D}_{\text{tr}}}$} is linearly independent, and {\small$\operatorname{dim}\left(\operatorname{span}\left(\left\{\mathbb{E}_{\hat{G}_{i}}\left[\rho(\hat{G})_i \rho(\hat{G})_i^{ \top}\right] \left(\Theta_{\phi}^\top \Theta_\omega - \Theta_{\tilde{\psi}}^\top \Theta_\gamma\right) \right\}_{i\in \mathcal{V}}\right)\right)>\operatorname{dim}(\phi) -r$}, then,
%\begin{equation*}	
$$
\operatorname{dim}\left(\operatorname{span}\left(\left\{\mathbb{E}_{\rho(\hat{G}^{D})}[\rho(\hat{G}^{D})\rho(\hat{G}^{D})^{\top}] \left(\Theta_{\phi}^\top \Theta_\omega - \Theta_{\tilde{\psi}}^\top \Theta_\gamma\right) -\mathbb{E}_{\rho(\hat{G}^{D}),\epsilon^D}[\rho(\hat{G}^{D})\epsilon^D]\right\}_{D\in \mathcal{D}_{\text{tr}}}\right)\right)>\operatorname{dim}(\phi) -r
$$
%\end{equation*}

We examine the two component individually.
Suppose that
\begin{equation}
   \operatorname{dim}(\operatorname{span}\left\{\mathbb{E}_{\rho(\hat{G}^{D})}[\rho(\hat{G}^{D})\rho(\hat{G}^{D})^{\top}] \left(\Theta_{\phi}^\top \Theta_\omega - \Theta_{\tilde{\psi}}^\top \Theta_\gamma\right)\right\}_{D\in \mathcal{D}_{\text{tr}}} = k.
\end{equation}
Since the set { $\left\{\mathbb{E}_{\rho(\hat{G}^{D})}[\rho(\hat{G}^{D})\rho(\hat{G}^{D})^{\top}] \left(\Theta_{\phi}^\top \Theta_\omega - \Theta_{\tilde{\psi}}^\top \Theta_\gamma\right)\right\}_{D\in \mathcal{D}_{\text{tr}}}$} is linearly independent,
and \\
{$\operatorname{dim}\left(\operatorname{span}\left(\left\{\mathbb{E}_{\hat{G}_{i}}\left[\rho(\hat{G})_i \rho(\hat{G})_i^{ \top}\right] \left(\Theta_{\phi}^\top \Theta_\omega - \Theta_{\tilde{\psi}}^\top \Theta_\gamma\right) \right\}_{i\in \mathcal{V}}\right)\right)>\operatorname{dim}(\phi) -r$},
we have $k > \operatorname{dim}(\phi) -r$.

Next, we consider $\mathbb{E}_{\rho(\hat{G}^{D}),\epsilon^D}[\rho(\hat{G}^{D})\epsilon^D]$. 
Since $\operatorname{rank} (A) \geq \operatorname{rank}(A B) $, and both $\epsilon^D$ and $\left(\Theta_{\phi}^\top \Theta_\omega - \Theta_{\tilde{\psi}}^\top \Theta_\gamma\right)$ are scalar values that do not affect the dimension,
we have 
\begin{equation}
\begin{aligned}
	&\operatorname{dim}(\operatorname{span}\left(\left[\mathbb{E}_{\rho(\hat{G}^{D}),\epsilon^D}[\rho(\hat{G}^{D})\epsilon^D]\right]\right) \\
	\geq & \operatorname{dim}(\operatorname{span}\left\{\mathbb{E}_{\rho(\hat{G}^{D})}[\rho(\hat{G}^{D})\rho(\hat{G}^{D})^{\top}] \left(\Theta_{\phi}^\top \Theta_\omega - \Theta_{\tilde{\psi}}^\top \Theta_\gamma\right)\right\}_{D\in \mathcal{D}_{\text{tr}}} = k.
\end{aligned}
\end{equation}

Taking the dimensions of both components into account, we arrive at
\begin{equation}
%\small
\begin{aligned}
&\operatorname{dim}\left(\operatorname{span}\left(\{\mathbb{E}_{\rho(\hat{G}^{D})}[\rho(\hat{G}^{D})\rho(\hat{G}^{D})^{\top}] \left(\Theta_{\phi}^\top \Theta_\omega - \Theta_{\tilde{\psi}}^\top \Theta_\gamma\right) 
-\mathbb{E}_{\rho(\hat{G}^{D}),\epsilon^D}[\rho(\hat{G}^{D})\epsilon^D]\}_{D\in \mathcal{D}_{\text{tr}}}\right)\right) \geq  k > \operatorname{dim}(\phi) -r.
\end{aligned}
\end{equation}

\textbf{Step 3:} We prove that if $\Theta_{\phi}^\top \Theta_\omega$ satisfies:  $\Theta_{\phi} \mathbb{E}_{\rho(\hat{G}^{D})}[\rho(\hat{G}^{D}){\rho(\hat{G}^{D})^{\top}}]\Theta_{\phi}^\top \Theta_\omega=\Theta_{\phi} \mathbb{E}_{\rho(\hat{G}^{D}),{Y}^{D}}[\rho(\hat{G}^{D}){Y}^{D}]$, for all $D\in \mathcal{D}_{tr}$ and  {\small$\operatorname{dim}(\operatorname{span}(\{\mathbb{E}_{\rho(\hat{G}^{D})}[\rho(\hat{G}^{D})\rho(\hat{G}^{D})^{\top}] \left(\Theta_{\phi}^\top \Theta_\omega - \Theta_{\tilde{\psi}}^\top \Theta_\gamma\right) -\mathbb{E}_{\rho(\hat{G}^{D}),\epsilon^D}[\rho(\hat{G}^{D})\epsilon^D]\}_{D\in \mathcal{D}_{\text{tr}}}))>\operatorname{dim}(\phi) -r$},
then $\Theta_{\phi}^\top \Theta_\omega=\Theta_{\tilde{\psi}}^\top \Theta_{\gamma}$ is causal invariant defender for all attack domain set $\mathcal{D}_{\text{all}}$,

According to $Y=C \gamma+\epsilon$, $\tilde{\psi}(\rho(\hat{G}))=C$, and Step 1, we have 
\begin{equation}
\begin{aligned}
   &\Theta_{\phi} \mathbb{E}_{\rho(\hat{G}^{D})}\left[\rho(\hat{G}^{D}){\rho(\hat{G}^{D})^{\top}}\right]\Theta_{\phi}^\top \Theta_\omega \\
   =& \Theta_{\phi} \mathbb{E}_{\rho(\hat{G}^{D}),{Y}^{D}}\left[\rho(\hat{G}^{D}){Y}^{D}\right] \\
   =& \Theta_{\phi} \mathbb{E}_{\rho(\hat{G}^{D}),{\epsilon}^{D}}\left[\rho(\hat{G}^{D})\left(\left(\Theta_{\tilde{\psi}} \rho(\hat{G}^{D})\right)^{\top}  \Theta_\gamma+\epsilon^D\right)\right].
   \label{eq:s3}
\end{aligned}
\end{equation}
We can re-write the Eq.~\ref{eq:s3} as:
\begin{equation}
    \Theta_{\phi} \left(\underbrace{ \mathbb{E}_{\rho(\hat{G}^{D})}\left[\rho(\hat{G}^{D}){\rho(\hat{G}^{D})^{\top}}\right]\Theta_{\phi}^\top \Theta_\omega - \mathbb{E}_{\rho(\hat{G}^{D}),{\epsilon}^{D}}\left[\rho(\hat{G}^{D})\left(\left(\Theta_{\tilde{\psi}} \rho(\hat{G}^{D})\right)^{\top}  \Theta_\gamma+\epsilon^D\right)\right]} _{:=t_D} \right) = 0
\end{equation}

To show that $\Phi$ leads to the desired invariant defender $\Theta_{\phi}^\top \Theta_\omega = \Theta_{\tilde{\psi}}^\top \Theta_\gamma$,
we assume $\Theta_{\phi}^\top \Theta_\omega \neq \Theta_{\tilde{\psi}}^\top \Theta_\gamma$ and reach a contradiction. First, according to Step 2, we have {\small$\operatorname{dim}(\operatorname{span}(\{\mathbb{E}_{\rho(\hat{G}^{D})}[\rho(\hat{G}^{D})\rho(\hat{G}^{D})^{\top}] \left(\Theta_{\phi}^\top \Theta_\omega - \Theta_{\tilde{\psi}}^\top \Theta_\gamma\right) -\mathbb{E}_{\rho(\hat{G}^{D}),\epsilon^D}[\rho(\hat{G}^{D})\epsilon^D]\}_{D\in \mathcal{D}_{\text{tr}}}))>\operatorname{dim}(\phi) -r$}.
Second, according to Step 1, each $t_D\in \operatorname{Ker}(\phi)$. 
Therefore, it would follow that $\operatorname{dim}(\operatorname{Ker}(\Theta_\phi))> \operatorname{dim}(\Theta_\phi) -r$, which contradicts the assumption that $\operatorname{rank}(\Theta_\phi) =r$, which is similar to~\cite{arjovsky2019invariant}.
Therefore, $\Phi$ leads to the desired invariant defender $\Theta_{\phi}^\top \Theta_\omega = \Theta_{\tilde{\psi}}^\top \Theta_\gamma$.

% This concludes the proof of the proposition.
\end{proof}

\clearpage
\section{Algorithm}
\label{apd:algo}
% In this section, we describe the training process for IDEA in Algorithm~\ref{alg:all}. We first optimize the model $f$ using Algorithm~\ref{alg:idea}. Specifically, we compute the representation $z$ of minibatch nodes $V_t$ by encoder $h$ and the attack domain $D$ by domain learner $s$. We then obtain the predictions $\hat{y}$ and $\hat{y_d}$ by classifiers $g$ and $g_d$, and compute the total loss. After updating $f$, we optimize the attack method and the domain learner. This procedure is repeated the number of training iterations. 

In this section, we present the training process for the IDEA algorithm, as illustrated in Algorithm~\ref{alg:all}. The model $f$ is first optimized using Algorithm~\ref{alg:idea}. During this optimization, the encoder $h$ calculates the representation $z$ for a minibatch of nodes $V_t$, and the domain learner $s$ identifies the attack domain $D$. Next, classifiers $g$ and $g_d$ produce predictions $\hat{y}$ and $\hat{y_d}$, respectively, which are then used to compute the total loss. After updating $f$, both the attack method and domain learner are optimized. This procedure is repeated iteratively for the number of training iterations.
% For each iteration, we first train the discriminator for several steps, and then train the attack generator once, following ~\cite{goodfellow2014generative}.
 \vspace{15pt}

\begin{algorithm}
\caption{The training process for IDEA method}
\begin{algorithmic}[1]
\label{alg:all}
\REQUIRE  clean graph $G=(\mathcal{V}, \mathcal{E}, X)$, attack method $\Lambda$,  set of node labels $Y$
\ENSURE  model $f$ concluding encoder $h$, classifiers $g$ and $g_d$, domain learner $s$
\FOR{number of training iterations}
%     \STATE{Sample minibatch of nodes $V_t$ from node set $\mathcal{V}$,  $V_t =\text{Sample}(V)$}
%   \STATE{Sample a neighbor for each $v$ in $V_t$ and obtain neighbor nodes $\mathcal{N}_t$,  $\mathcal{N}_t=\text{NeighbSample}(V_t)$}
%   \STATE{Generate the perturbed graph by attack method $\Lambda$,  $\hat{G}=\Lambda(G)$}
%   \STATE{Compute the representation by the encoder $h$ on clean graph $G$ (i.e., $z_\text{cln}$) and on perturbed graph $\hat{G}$ (i.e., $z_{\tex{ptb}}$),  $z_\text{cln}=h(G)[V_t]$ , $z_\text{ptb}=h(\hat{G})[V_t]$}
%   \STATE{Obtain the total representation $z$ by concatenating $z_\text{cln}$ and $z_\text{ptb}$,  $z=\text{Concat}(z_\text{cln}, z_\text{ptb})$}
%   \STATE{Obatin the attack domain $D$ by domain learner $s$,  $D=s(z)$}
%   \STATE{Compute the prediction $\hat{y}$  and prediction based on attack domain $\hat{y_a}$ for nodes $V_t$ by the classifier $g$ and $g_{d}$,  $\hat{y} = g(z)$,  $\hat{y_d} = g_d(z,D)$ }
%   \STATE{Compute the predictive loss $\mathcal{L}_\mathcal{P}$, node-based invariance loss $\mathcal{L}_\mathcal{I}$, and structural-based invariance loss $\mathcal{L}_\mathcal{E}$ by Eq.5, Eq.8, and Eq.9, respectively.}
%   \STATE{Compute the total loss $\mathcal{L}=\mathcal{L}_\mathcal{P} + \alpha\mathcal{L}_\mathcal{I} + \beta\mathcal{L}_\mathcal{E}$}
%   \STATE{Compute the gradient of the model $f$ and update $f$.
 \STATE{{Sample minibatch of nodes $V_t$ from node set $\mathcal{V}$,  $V_t =\text{Sample}(V)$}
 \\ \textit{\% Optimize the model $f$}}
 \STATE{Update the model $f$ by Algorithm~\ref{alg:idea}}
 \STATE{{Sample minibatch of nodes $V_t$ from node set $\mathcal{V}$,  $V_t =\text{Sample}(V)$}
 \\ \textit{\% Optimize the attack method}}
 \STATE{Generate the perturbed graph by attack method $\Lambda$,  $\hat{G}=\Lambda(G)$}
 \STATE{Compute the prediction $\hat{y}$ by the classifier $g$,  $\hat{y} = g(z_\text{atk})$, where $z_\text{atk} =h(\hat{G})[V_t]$}
 \STATE{Compute the attack loss $\mathcal{L}_\text{atk}=-\mathcal{L}_\mathcal{P}$ , where $\mathcal{L}_\mathcal{P}$ is computed by Eq.5}
 \STATE{Compute the gradient of attack method $\Lambda$ and update $\Lambda$. }
 \STATE{Sample minibatch of nodes $V_t$ from node set $\mathcal{V}$,  $V_t =\text{Sample}(V)$
 \\ \textit{\% Optimize the domain learner }}
 \STATE{Generate the perturbed graph by attack method $\Lambda$,  $\hat{G}=\Lambda(G)$}
 \STATE{Obtain  total representation $z$ by concatenating $z_\text{cln}$ and $z_\text{ptb}$,  $z=\text{Concat}(z_\text{cln}, z_\text{ptb})$, where $z_\text{cln}=h(G)[V_t]$ , $z_\text{ptb}=h(\hat{G})[V_t]$}
 \STATE{Obatin the attack domain $D$ by domain learner $s$,  $D=s(z)$}
 \STATE{Compute the prediction $\hat{y}$ by the classifier $g$,  $\hat{y} = g(z)$}
 \STATE{Compute the loss for the domain learner $s$ by Eq.11}
 \STATE{Compute the gradient of domain learner $s$ and update $s$ }
 \ENDFOR
\end{algorithmic}

\end{algorithm}

\begin{algorithm}[b]
\caption{The algorithm of IDEA}
\label{alg:idea}
\begin{algorithmic}[1]
\REQUIRE  clean graph $G=(\mathcal{V}, \mathcal{E}, X)$, attack method $\Lambda$,  set of node labels $Y$,  minibatch nodes $V_t$
\ENSURE updated model $f$
 \STATE{Sample a neighbor for each $v$ in $V_t$ and obtain neighbor nodes $\mathcal{N}_t$,  $\mathcal{N}_t=\text{NeighbSample}(V_t)$}
 \STATE{Generate the perturbed graph by attack method $\Lambda$,  $\hat{G}=\Lambda(G)$}
 \STATE{Compute the representation by the encoder $h$ on clean graph $G$ (i.e., $z_\text{cln}$) and on perturbed graph 
 $\hat{G}$ (i.e., $z_{\text{ptb}}$),  $z_\text{cln}=h(G)[V_t]$, $z_\text{ptb}=h(\hat{G})[V_t]$}
 \STATE{Obtain the total representation $z$ by concatenating $z_\text{cln}$ and $z_\text{ptb}$,  $z=\text{Concat}(z_\text{cln}, z_\text{ptb})$}
 \STATE{Obatin the attack domain $D$ by domain learner $s$,  $D=s(z)$}
 \STATE{Compute the prediction $\hat{y}$  and prediction based on attack domain $\hat{y_d}$ for nodes $V_t$ by the classifier $g$ and $g_{d}$,  $\hat{y} = g(z)$,  $\hat{y_d} = g_d(z,D)$ }
 \STATE{Compute the predictive loss $\mathcal{L}_\mathcal{P}$, node-based invariance loss $\mathcal{L}_\mathcal{I}$, and structural-based invariance loss $\mathcal{L}_\mathcal{E}$ by Eq.5, Eq.8, and Eq.9, respectively.}
 \STATE{Compute the total loss $\mathcal{L}=\mathcal{L}_\mathcal{P} + \mathcal{L}_\mathcal{I} + \mathcal{L}_\mathcal{E}$}
 \STATE{Compute the gradient of the model $f$ and update $f$}
 \STATE{\textbf{Return} model $f$}
\end{algorithmic}
\end{algorithm}

%% Biography
%% Here goes the biography details.

%% Here goes the biography details.

\end{document}